\documentclass{article}

\usepackage[final]{neurips_2023}

\usepackage[utf8]{inputenc} % allow utf-8 input
\usepackage[T1]{fontenc}    % use 8-bit T1 fonts
\usepackage{hyperref}       % hyperlinks
\usepackage{url}            % simple URL typesetting
\usepackage{booktabs}       % professional-quality tables
\usepackage{amsfonts}       % blackboard math symbols
\usepackage{nicefrac}       % compact symbols for 1/2, etc.
\usepackage{microtype}      % microtypography
\usepackage{xcolor}         % colors
\usepackage{enumitem}
\usepackage{amsmath}

\newcommand{\taskname}{\textsc{D5}}
\newcommand{\datasetname}{\textsc{Open\taskname}}
\newcommand{\syndataset}{\textsc{Syn\taskname}}

\usepackage{tikz}
\newcommand*\circled[1]{\textcircled{\raisebox{-0.9pt}{#1}}}

\title{Goal Driven Discovery of Distributional Differences via Language Descriptions}

\author{%
  Ruiqi Zhong\thanks{University of California, Berkeley, EECS Department. Email: ruiqi-zhong@berkeley.edu}, Peter Zhang, Steve Li, Jinwoo Ahn, Dan Klein, Jacob Steinhardt}
\begin{document}

\maketitle

\begin{abstract}
    Exploring large corpora can generate useful discoveries but is time-consuming for humans.
    We formulate a new task, \taskname{}, that automatically discovers differences between two large corpora in a goal-driven way. 
    The task input is a problem comprising a user-specified exploration goal (``\textit{comparing the side effects of drug A and drug B}'') and a corpus pair (collections of patients' self-reported reactions after taking each drug). 
    The output is a goal-relevant description (discovery) of how these corpora differ (patients taking drug A ``\textit{mention feelings of paranoia}'' more often).
    We build a \taskname{} system, and to quantitatively evaluate its performance, we 1) build a diagnostic benchmark, \syndataset, to test whether it can recover known differences between two synthetic corpora, and 2) contribute a meta-dataset, \datasetname{}, aggregating 675 open-ended problems ranging across business, social sciences, humanities, machine learning, and health.
    With both synthetic and real datasets, we confirm that language models can leverage user-specified goals to propose more relevant candidate discoveries, and they sometimes produce discoveries previously unknown to the authors, including demographic differences in discussion topics, political stances in speech, insights in commercial reviews, and error patterns in NLP models.
    Finally, we discuss the limitations of our \taskname{} system, which discovers correlation rather than causation and potentially reinforces biases when misused; therefore, practitioners should treat the outputs of our system with caution.
\end{abstract}

\section{Introduction}

Exploring large corpora and generating discoveries from them can be ad hoc and laborious.
For example, to compare the side effects of drug A and drug B, doctors might inspect two large corpora of patients' self-reported reactions after taking each drug;
based on ad hoc insights, they hypothesize that patients taking drug A more often ``\textit{mentions feelings of paranoia}'', and then validate this hypothesis by laboriously inspecting the two corpora.  
Since machines can automatically process a large amount of texts, we might hope for ML systems to facilitate exploratory analyses like the one above.

\begin{figure*}[t!]
    \centering
    \includegraphics[width=\linewidth]{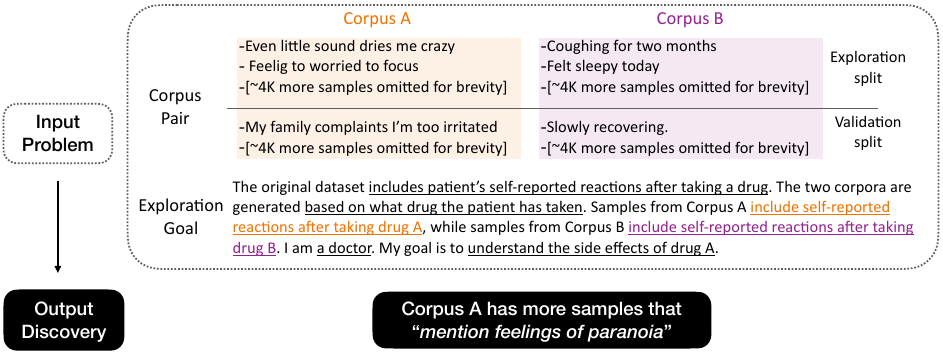}
    \caption{Each problem in \datasetname{} contains 1) a corpus pair, which has $\sim$17K samples on average and is partitioned into two halves called ``exploration split'' and ``validation split'', and 2) a natural language description of the exploration goal, which also contains information about how the corpus pair was collected. 
    A \taskname{} system takes the goal and the exploration split as inputs and generates valid and relevant discoveries in natural language as outputs.
    The underlined texts in the exploration goal vary across problems, while the rest are templates. 
    }
    \label{fig:task-description}
\end{figure*}

However, an ML task requires a unified input-output space and evaluation metric so that it can be automated, benchmarked, learned, and analyzed.
To this end, we formalize one type of exploratory analysis problem as a natural language generation task: goal \textbf{d}riven \textbf{d}iscovery of \textbf{d}ifferences between text \textbf{d}istributions via language \textbf{d}escriptions (\taskname{}).
As shown in Figure \ref{fig:task-description}, the input to the \taskname{} task is a ``problem'' comprising a description of a user-specified exploration goal (understanding side effects) and a corpus pair (text samples from the distributions of self-reported reactions after taking each drug). 
The output is a ``discovery'' represented as a natural language predicate (``\textit{mentions feelings of paranoia}'').
We evaluate a discovery with two criteria (Section~\ref{sec:metrics}):
(1) validity: it should describe a true difference \citep{zhong2022describing}; and
(2) relevance to the goal \citep{mcgarry2005survey}.

%However, fully evaluating a \taskname{} system is challenging. 
%Since \taskname{} is most useful when it discovers differences humans were unaware of, the most popular benchmark practice -- comparing system-generated outputs with human-written references on a set of representative problems -- is infeasible.
%Even worse, fully validating the model-generated discoveries is labor intensive (Section \ref{sec:application}).
Since \taskname{} is open-ended and aims at discovering unknowns, the most popular benchmark practice---comparing system-generated outputs with human-written references on a test set---is infeasible.
We therefore design two evaluation strategies. 
\begin{itemize}
[topsep=0pt,itemsep=-1ex,partopsep=1ex,parsep=1ex,leftmargin=2em]
    \item \textbf{Diagnostic}: we synthesized a dataset of \taskname{} problems with known solutions, \syndataset, to diagnose whether a \taskname{} system can recover known differences between two synthetic corpora. This strategy is cheap and automated but might not reflect user utility in real applications.
    \item \textbf{Open-ended}: we collected a dataset, \datasetname{}, by aggregating 675 {open}-ended \taskname{} problems ranging across business, social sciences,  humanities, health, and machine learning (Figure \ref{fig:dataset-overview}), comprising 4.4 million text samples in total across problem corpora. We then manually evaluated a subset of the output discoveries. This strategy is subjective and expensive, but useful for obtaining qualitative insights on more realistic applications.
\end{itemize}

% \begin{itemize}
%   \item Diagnostic: we synthesized a dataset of \taskname{} problems, \syndataset, to diagnose whether a \taskname{} system can recover known differences between two synthetic corpora. 
%   %The corpora contain texts with different properties, such as topics, styles, and languages, and the goal is to identify property-specific differences. 
%   This diagnostic setup is cheap and automated, but might not reflect user utility on real applications.
%   \item Open-ended: we collected a metadataset, \datasetname{}, by aggregating 675 {open}-ended \taskname{} problems ranging across business, social sciences,  humanities, health, and machine learning (Figure \ref{fig:dataset-overview}), comprising 4.4 million text samples in total across problem corpora. We collected these 675 problems by surveying papers that focus on text analysis (e.g.~\citet{nguyen2020we}), brainstorming exploration goals, scraping the corresponding corpora, and post-processing them over nine months. 
%   We then manually evaluated a subset of the output discoveries. 
%   This setup is subjective and expensive, but useful for obtaining qualitative insights on more realistic applications.
% \end{itemize}

These two strategies allow us to quantitatively evaluate and compare \taskname{} systems. 
For example, we compared 1) the system from \citet{zhong2022describing} designed to describe corpus-level differences without goals, and 2) a goal-conditioned variant that we develop in Section \ref{sec:method}.
We found language models successfully use the specified goal: the goal-conditioned variant is correct 12\% more often on \syndataset{}, and it produces relevant candidate discoveries 31\% more often on \datasetname{}.

%Our system closely mirrors that of \citet{zhong2022describing}, which also describes differences in text corpora, but which does not consider the goal for either modeling or evaluation.

We envision \datasetname{} to be a growing, diverse repository of open-ended \taskname{} problems.
They will not only help us evaluate \taskname{} systems more reliably, but also allow the following operations:

\noindent\textbf{Facilitate exploratory analysis.}
Every time we build a better \taskname{} system, we can apply it to a repository of open problems and send the discoveries to researchers who posed them.
We show this paradigm is plausible by using our system to automatically produce useful discoveries on \datasetname{} (Section \ref{sec:application}), including insights from commercial reviews, temporal and demographic differences in discussion topics, political stances and stereotypes in speeches, differences in lyric styles, and error patterns in NLP systems. 
We anticipate future systems to produce more discoveries.

\noindent\textbf{Analyze the limitations of our evaluation.} 
Using concrete examples from \datasetname{}, we show that our current evaluation metrics do not encourage diverse findings, do not always produce causal conclusions, and cannot evaluate discoveries involving heavy expert knowledge (Section~\ref{sec:analysis}).
More \taskname{} problems can help us identify more limitations, which inform areas for future improvement. 

\noindent\textbf{Train better \taskname{} systems.}
Like other ML tasks, we can train a system once we have a dataset.
We describe a self-supervised learning algorithm that uses a repository of problems (without reference solutions) to train LMs to propose more valid hypotheses (Section \ref{sec:training}).
As a proof-of-concept, we show that it can make LMs better describe the differences between small groups of text samples.

To conclude, we show that \taskname{} can be quantitatively evaluated, automated, analyzed, and learned. 
Like other ML tasks, it would benefit from a more diverse, authentic, and larger dataset. 
We hope future works can gather feedback from domain experts and curate an ever-larger dataset of \taskname{} problems, thus accelerating exploratory analyses and facilitating scientific discoveries.
\footnote{Our code is released at \url{https://github.com/ruiqi-zhong/D5} and our code to download \datasetname{} is released at \url{https://github.com/petezh/OpenD5}. Given the limitations of our system, practitioners should interpret its outputs with caution and not use it to fully automate scientific discoveries.}
%\footnote{We share the code in our supplementary material and the \datasetname{} dataset at \url{https://doi.org/10.5281/zenodo.7683302}. The license information is in Appendix \ref{app:dataset-descriptions}.}

\section{Datasets: \syndataset{} and \datasetname{} } \label{sec:data}

We first introduce how each input problem is formatted. 
Then we discuss 1) how we synthesized \syndataset, which is used for automatic diagnostic evaluation, and 2) how we collected \datasetname{}, which is used to investigate the practical value of \taskname{} systems in open-ended applications.

\subsection{Task Format}

Each \taskname{} problem is represented by a corpus pair (Corpus A/B) and a description of the exploration goal.
For example,   Corpus A/B might be self-reported reactions after taking drug A/B, and the goal description would be ``\textit{comparing the side effects of drug A and drug B}''.
The desired output is valid and relevant discoveries in the form of natural language predicates (Figure \ref{fig:task-description}), e.g. Corpus A has more samples that ``\textit{mentions feelings of paranoia}''.

\subsection{\syndataset{}, a Diagnostic Benchmark with Reference Solutions and an Automatic Metric} \label{sec:synd5}

To automatically diagnose a \taskname{} system, we synthesized \syndataset{}, a dataset of \taskname{} problem with reference solutions.
To synthesize Corpus A and Corpus B for each input problem, we used a language model (LM) to generate two corpora that simultaneously differ on two dimensions, one of which is goal-relevant and one of which is a distractor. For instance, suppose the goal is to ``\textit{understand how Corpus A differs from Corpus B in terms of topic}''. Then we would synthesize an example where Corpus A is more sports-related while B is more art-related (goal-relevant: varying topic), while additionally Corpus A is in English while B is in French (distractor: varying language). 
The reference solution is the difference on the goal-relevant dimension, e.g. ``\textit{is sports-related}''.

In more detail, to synthesize an example in \syndataset{}, we first picked one goal-relevant and one distractor dimension from the set $\{\textrm{topic}, \textrm{genre}, \textrm{language}\}$, and sampled a value for each corpus and dimension (e.g. Corpus A: [sports, English]; Corpus B: [art, French]).
We then synthesized Corpus A/B such that all its samples are in English/French (i.e. completely different on the distractor dimension) while $V$ percent of them are sports-related/art-related, where we varied $V$ from 0.6 to 1.
Since the distractor difference is more salient, \syndataset{} penalizes \taskname{} systems that ignore the goal and output the incorrect distractor difference ``\textit{is in English}''.
We synthesized 300 problems in total to create \syndataset{}; see Appendix \ref{app:synd5} for a detailed description of the pipeline.

To compute a \taskname{} system's accuracy, we prompted \texttt{Claude-v1.3} \citep{bai2022constitutional} to judge how often the output discovery is semantically equivalent to the reference.
We construct the prompt by using 6 pairs of predicates with the labels of ``equivalent'', ``similar'', or ``irrelevant'' as few-shot examples, and ask \texttt{Claude-v1.3} to judge whether the output discovery and the reference are ``equivalent''.
As a result, we can automatically diagnose a \taskname{} system.
See Appendix \ref{app:similarity} for the prompt for equivalence judgement and Appendix \ref{app:robustness-check-similarity} for two robustness checks, which (a) consider ``similar'' discoveries to be correct as well, and (b) use other LMs for equivalence judgement.

\begin{figure*}[t!]
    \centering
    \includegraphics[width=\linewidth]{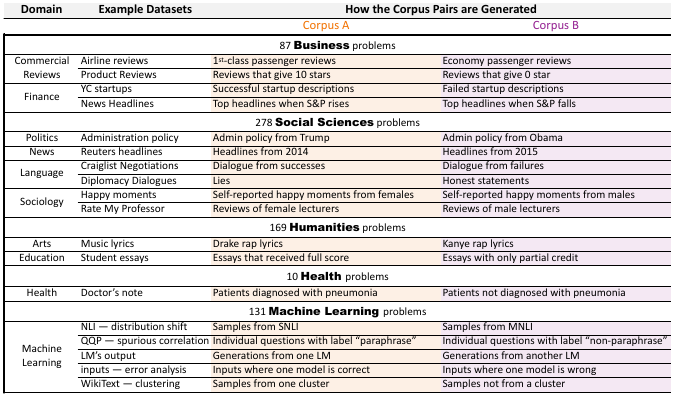}
    \caption{
    \datasetname{} contains 675 problems. See citations in Appendix \ref{app:dataset-descriptions}.
    }
    \label{fig:dataset-overview}
\end{figure*}

\subsection{\datasetname{}, a Realistic Open-Ended Dataset without Reference Solutions}
To evaluate a \taskname{} system's utility under realistic applications, we also gathered \datasetname{}, a realistic dataset of 675 open-ended \taskname{} problems.
These problems range across business, social sciences, humanities, health, and machine learning; see Figure \ref{fig:dataset-overview} for a few examples.
To build \datasetname, two of the authors performed an extensive literature review on problems that could potentially benefit from our system, e.g., reading survey papers \citep{nguyen2020we} and courses on computational social sciences, and skimming through the ACL proceedings from the past decade and datasets from Kaggle that have an NLP tag; we then annotated the exploration goals, scraped/generated the corresponding corpora, and post-processed them over nine months (see complete list of citations in Appendix \ref{app:dataset-descriptions}).
As shown in Figure \ref{fig:task-description}, each goal describes the original dataset, how the two given corpora are generated, who is using the system, and what property of the two corpora does the user want to understand.
Each \datasetname{} problem is reviewed by at least two of the authors to reduce grammatical mistakes and ambiguous interpretations of the goal.

Each corpus contains around 17K text samples on average, and \datasetname{} in total comprises 4.4 million distinct text samples.
We use 50\% of each corpus as the ``exploration'' split and 50\% as the ``validation'' split.
The system can only access the exploration split, while the validation split is reserved for the evaluators to validate the discovery. 
A validation split prevents overfitting the discoveries to the given samples and is analogous to the train-test split in machine learning.

Since we hope to build systems that can tackle challenging open-ended problems, we did not avoid cases where we do not know the ground truth answer. 
% This means that, for some problems, it might be infeasible to produce any meaningful discovery. 
This is different from standard benchmarking practices, where humans can provide a reference solution to evaluate an AI system.
However, even though we do not know the ground truth, once a system produces a discovery, we can still evaluate it. 
We present our evaluation metrics in the next section.

\section{Evaluation Metrics for Open-Ended \taskname{} problems} \label{sec:metrics}

For the goal of comparing the side effects of drug A and drug B, how do we evaluate a system-generated discovery that Corpus A ``\textit{mention feelings of paranoia}'' more often?
First, it needs to be valid, such that indeed more samples from Corpus A satisfy this predicate, which can be evaluated (approximately) objectively.
Second, it needs to be relevant to the goal of understanding side effects, which depends on the user's subjective judgement.
We define validity and relevance below.

\noindent \textbf{Validity.}
Similar to \citet{zhong2022describing}, we require an output discovery $h$ to be a truth predicate on a text sample. 
For example, if $h$ = ``\textit{mentions about family}'', then $h$ is true on the string $x_{1}$ = ``\textit{My daughter loves me}'' and false on the string $x_{2}$ = ``\textit{I'm going to school}''.
Define $T(h, x) \in [0, 1]$ as ``the certainty that $h$ is true on $x$'', e.g., $T(h, x_{1}) \approx 1$ and $T(h, x_{2}) \approx 0$.
We approximate $T(h, x)$ by asking three Turkers how certain they are and averaging their responses (see Appendix \ref{app:turker} for details).

Let $\mathcal{D}^{\text{val}}_{A}$ and $\mathcal{D}^{\text{val}}_{B}$ denote the validation sets for Corpus A and B. We define the ``validity'' $V$ as
\begin{equation} \label{eq:soundness}
    V(h) := \mathbb{E}_{x\sim \mathcal{D}^{\text{val}}_{A}}[T(h, x)] -  \mathbb{E}_{x\sim \mathcal{D}^{\text{val}}_{B}}[T(h, x)].
\end{equation}
Computing $V(h)$ is expensive since it requires human annotations $T(h, x)$ on a set of text samples even to evaluate a single discovery $h$. 
In practice, we do not have the budget to compute $V(h)$ on the entire validation split;
therefore, we approximate this quantity by randomly sampling from Corpus $A$ and Corpus $B$.
We use these samples to compute an empirical estimate of $V$, as well as a $p$-value for the null hypothesis that $V \leq 0$ using a one-sided t-test. 
% In this paper, we call a discovery valid if its $p$-value is less than 10\%.

\noindent \textbf{Relevance.}
A discovery may be irrelevant even if $V=1$.
For example, if the goal is to understand the writing style differences between higher-scoring essays (Corpus A) and lower-scoring ones (Corpus B), the discovery that Corpus A ``\textit{achieves higher scores}'' has high validity score by definition but irrelevant to the goal of understanding stylistic differences.

Therefore, we designed a procedure to evaluate relevance, where human or language model evaluators would score each discovery with \circled{2}/\circled{1}/\circled{0}.
The evaluators used the rubric below, which illustrates the meaning of each score with the essay example above:
\begin{itemize}[topsep=0pt,itemsep=-1ex,partopsep=1ex,parsep=1ex,leftmargin=2em]
    \item \circled{2}, relevant; e.g. the discovery ``\textit{write in first person}'' is directly related to the writing style.
    \item \circled{1}, indirectly relevant; e.g. the discovery ``\textit{use the word “I”}'', is not exactly a writing style, but can still inform the relevant underlying principle of ``\textit{write in first person}''.
    \item \circled{0}, irrelevant; e.g. the discovery ``\textit{argue for abortion}'' is unrelated to the writing style. 
\end{itemize}

To minimize biases while comparing two systems, the evaluators are blind to which system generates which discoveries.

To conclude, an ideal discovery would have a high $V$ value with a small $p$-value and achieve ratings of \circled{2} in relevance.
In the next section, we will build a \taskname{} system that addresses these criteria by first proposing goal-relevant candidate discoveries (hypotheses) and then automatically validate them.

\noindent \textbf{Other metrics.} We also explored two other subjective metrics, novelty (how difficult it is to generate the discovery) and significance (how beneficial it is to learn about the discovery). 
Due to space limit, we present their rubrics and related results in Appendix \ref{app:more-metrics}.

\section{Methods: Building a \taskname{} System} \label{sec:method}

We describe our \taskname{} system, which maps from a corpus pair and an exploration goal to a set of natural language predicates.
Our system is inspired by a two-stage model of how humans discover patterns in data: creatively brainstorming hypotheses and then rigorously validating them on the data \citep{ludwig2022algorithmic}.
Analogously, we first propose hypotheses conditioned on the exploration goal and a subset of samples from the corpus pair (Section \ref{sec:proposer}). We then use a language model to approximately compute the validity of each hypothesis, and output the valid ones as the final discoveries (Section \ref{sec:verifier}). 
Our system closely mirrors that of \citet{zhong2022describing}, except that we leverage the goal to propose more relevant hypotheses.
Finally, we present a self-supervised learning algorithm to improve an LM's ability to propose more valid hypotheses (Section \ref{sec:training});
however, due to API access constraint, we cannot apply it to fine-tune \texttt{gpt-3}, so we provide a proof of concept experiment on \texttt{Flan-T5} \citep{chung2022scaling}.

\begin{figure*}[h!]
    \centering
    \includegraphics[width=0.9\linewidth]{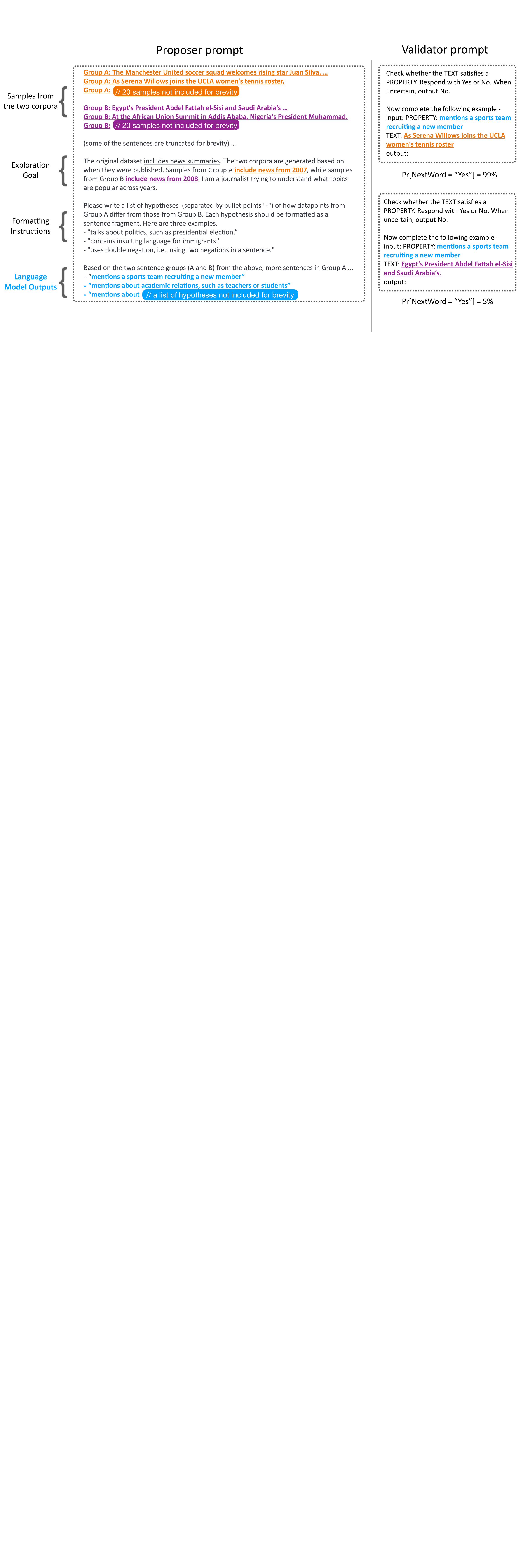}
    \caption{All underlined content in the prompt differs across problems, while the other content in the prompt is templated. 
    \textbf{Left}: proposer prompt. The generated hypotheses are in blue. All content with colored background is excluded for brevity. For the baseline of not using the exploration goal, we removed the ``exploration goal'' block from the prompt.
    \textbf{Right}: the validator prompt.}
    \label{fig:prompts}
\end{figure*}

\subsection{Hypothesis Proposer} \label{sec:proposer}
We prompt \texttt{gpt-3} \citep{ouyang2022training} to propose hypotheses.
Denoting the exploration split of Corpus A/B as $\mathcal{D}^{\text{exp}}_{A}$/$\mathcal{D}^{\text{exp}}_{B}$, we construct the prompt by concatenating a few random samples from $\mathcal{D}^{\text{exp}}_{A}$ and $\mathcal{D}^{\text{exp}}_{B}$, the exploration goal, and an instruction to output a list of hypotheses. Figure \ref{fig:prompts} (left) depicts an example of the resulting prompt, together with a typical language model output.

Since the entire corpus pair might not fit into one prompt, we construct multiple prompts with different sets of samples so that \texttt{gpt-3} can ``see'' as many different samples as possible in our pipeline.
We continue sampling hypotheses with different prompts until obtaining a set of 60 hypotheses, which we call $H_{\text{init}}$.
Appendix \ref{app:proposer} includes more details on selecting the sets of samples for different prompts. 
%for additional details on how the samples are selected.

\subsection{Hypothesis Validator} \label{sec:verifier}
\begin{figure*}[h!]
    \centering
    \includegraphics[width=\linewidth]{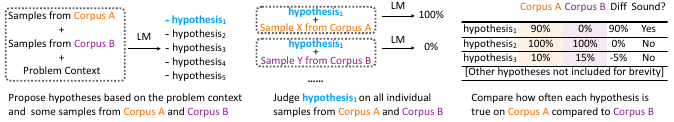}
    \caption{A sketch of the baseline method. The description can be seen in Section \ref{sec:method} and the actual prompts can be seen in Figure \ref{fig:prompts}. }
    \label{fig:baseline}
\end{figure*}

Many hypotheses in $H_{\text{init}}$ have low validity: they are not more often true on $\mathcal{D}_{A}$ than on $\mathcal{D}_{B}$ (i.e. $V(h) \leq 0$). 
To automatically filter them out, we use a language model $T'$ to simulate the Turkers' judgement $T$ and hence approximate the validity score $V(h)$ with the function %of a hypothesis $h$ with 
$V'(h)$, defined as 
\begin{equation}
        V'(h) := \mathbb{E}_{x\sim \mathcal{D}^{\text{exp}}_{A}}[T'(h, x)] - \mathbb{E}_{x\sim \mathcal{D}^{\text{exp}}_{B}}[T'(h, x)].
\end{equation}
To compute $T'$, we ask \texttt{Flan-T5} whether $x$ satisfies $h$ with the prompt shown in Figure \ref{fig:prompts} (right).
To better simulate Turker's judgment, we collected additional Turker annotations to fine-tune \texttt{FLAN-T5} (see Appendix~\ref{app:verifier} for details about the data collection process).
We then obtain a significance value $p'$ by performing a t-test to compare the mean value of $V'(h, x)$ on the exploration split of Corpus $A$ to that of Corpus $B$, rule out the hypotheses with $p'$ greater than 0.001, and output the remainder as discoveries.
Finally, we obtain additional discoveries by repeating the same process but asking our system to propose and validate hypotheses about Corpus $B$ rather than Corpus $A$.
Appendix Figure~\ref{fig:baseline} visualizes our entire pipeline and Appendix~\ref{app:compute} discusses the computational resources we used.

\subsection{Self-Supervised Learning with Open-Ended Problems: A Proof of Concept} \label{sec:training}
Since \taskname{} problems are open-ended, future systems could potentially produce discoveries with higher validity scores than any known discovery.
Therefore, we design a self-supervised learning algorithm to improve an LM's ability to propose more valid hypotheses, using the principle that \textbf{it is easier to validate a discovery than to generate one}.

\noindent\textbf{Algorithm.}
Suppose we are given a set of problems for training and an initial language model $m_{\text{0}}$.
Our goal is to automatically generate a set of \textit{prompt}-\textit{completion} pairs to fine-tune $m_{\text{0}}$ so that it can propose hypotheses that are more valid. 
To generate a \textit{prompt}, we randomly sample a problem and create a proposer prompt following the procedure in Section \ref{sec:proposer}.
To generate the desired \textit{completion} given a prompt, we sample multiple hypotheses from $m_{\text{init}}$, approximate their $V'$ score on the samples in the proposer prompt with the same language model $m_{\text{init}}$ (Section \ref{sec:verifier}), and select the highest scoring hypothesis.
Finally, we use the prompt-completion pairs to fine-tune $m_{\text{0}}$.

\noindent\textbf{A Proof of Concept Experiment.} Since we cannot fine-tune \texttt{text-davinci-003}, we can only experiment with \texttt{Flan-T5-xxl} ~\citep{chung2022scaling}, an open-sourced instruction-tuned model that might only work well for easier ``mini-problems''.
As a proof of concept, we tested the above self-supervised learning algorithm on the task of describing groups of four samples, where each group comes from a text cluster.

We computed both the automated ``self-evaluation'' validity score $V'$ and the ``true'' validity score $V$ according to Turker evaluation for evaluation.
After self-training, $V'$ improves substantially from 0.22 to 0.37, and the $V$ improves from 0.07 to 0.10, with a $p$-value of 0.02.
This result provides preliminary evidence that self-training could be applied to a large set of problems to improve the validity of the hypotheses; we expect future validators to simulate human judgments better, hence decreasing the approximated gap of improvement between $V$ and $V'$.
We discuss more training and evaluation detail in Appendix \ref{app:training}. 

\section{Quantitative Evaluation on \syndataset{} and \datasetname{}} \label{sec:quantitative}

We show that both \syndataset{} and \datasetname{} can be used to quantitatively evaluate \taskname{} systems.
Since \syndataset{} is automatic, we used it to compare a broad range of \taskname{} systems and studied the contributions of three different factors: the quality of the proposer model (\texttt{gpt-4} vs.~\texttt{text-davinci-003}), the use of a validator, and the use of a goal.
We then further investigated the effect of using goals under realistic applications through human evaluation on \datasetname{}.

\noindent \textbf{Automatically comparing different variants with \syndataset{}.}
As mentioned above, we ablated 3 factors, resulting in 2$^3$ = 8 variants. 
We compared 1) using \texttt{text-davinci-003} vs.~\texttt{gpt-4} as the hypothesis proposer; 2) using the validator to compute $V'$ for each hypothesis and outputting the highest-scoring hypothesis, vs.~not using the validator and outputting a random hypothesis; and 3) using the goal vs.~replacing it with ``\textit{I want to understand how Corpus A is different from Corpus B.}''.
We then automatically calculated the accuracy for each variant as described in Section \ref{sec:synd5}. 

We report the results in Table \ref{tab:synd5-table}. We find that using the validator and the goals significantly improve the performance, and \texttt{gpt-4} outperforms \texttt{text-davinci-003} with goals and the validator ($p$ < 1\% under a t-test).
We conducted two additional robustness checks in the Appendix \ref{app:robustness-check-similarity}: (a) using \texttt{text-davinci-003} instead of \texttt{Claude-v1.3} to judge predicate equivalence, and (b) considering discoveries semantically similar to the references also to be correct;
our conclusions do not change.

Finally, to improve the accessibility of our research, we ran the same experiments using \texttt{gpt-3.5-turbo} and \texttt{flan-t5-xxl} as our proposer, and report the results in Appendix Table \ref{tab:accessible}.
To show that our conclusions are general and not only apply to synthetically generated texts, we additionally constructed an extension of \syndataset{} with human-written texts by adapting the NYT dataset from \citet{wang2023goal}, where each text sample is a New York Times article with a topic and a location label: the topic dimension has 9 different values (e.g., politics, arts) and the location dimension has 10 different values (French, Italy);
we then followed the same procedure described in Section \ref{sec:synd5} to create this extension of \syndataset{}, and report our systems' performance in Appendix Table \ref{tab:nyt}.
In all experiments, using the validator and the goal improves the performance.

% We used \syndataset{} to automatically compare different variants of the system described in Section \ref{sec:method};
% specifically, we used an LM to compute how often the output discoveries with the highest $V'$ values are similar or equivalent to the reference solutions. 
% We ablated 3 factors, resulting in 2$^3$ = 8 variants:
% \begin{itemize}[topsep=0pt,itemsep=-1ex,partopsep=1ex,parsep=1ex,leftmargin=2em]
%     \item \textbf{Factor 1:} using \texttt{text-davinci-003} vs.~\texttt{gpt-4} as the hypothesis proposer;
%     \item \textbf{Factor 2:} w/ validator vs. wo/ validator. This aims to investigate whether the validator is indeed effective at ruling out less valid hypotheses. For variants without validator, we evaluated the average similarity across all proposed hypotheses.
%     \item \textbf{Factor 3:} w/ goal vs. wo/ goal. We investigated whether the proposer can leverage the goal successfully to produce relevant discoveries. For variants without the goal, we replaced the goal with the string ``\textit{I want to understand how Corpus A is different from Corpus B.}''
% \end{itemize}

\begin{table*}[t!]
    \centering
    \begin{tabular}{c|cc||c|ccc}
       \texttt{text-davinci-003}  & w/\phantom{o} goal & wo/ goal & \texttt{gpt-4} & w/\phantom{o} goal& wo/ goal\\
       \hline
       w/\phantom{o} validator  & 12\% & \phantom{0}2\% & w/\phantom{o} validator& 27\% & 15\% \\
       wo/ validator & \phantom{0}4\% & \phantom{0}1\% & wo/ validator&\phantom{0}8\% & \phantom{0}5\%\\
       \hline
    \end{tabular}
    \caption{The accuracy on \syndataset{} using different proposers, with/without incorporating goals, and with/without using validators. Using the validator, the goals, and \texttt{gpt-4} leads to better results.
    }
    \label{tab:synd5-table}
\end{table*}

\begin{table*}[t!]
\centering
    \begin{tabular}{l|ccc|c}
    
       Hypothesis Relevance  & \circled{2} & \circled{1} & \circled{0} & average \\
       \hline
       Using the goal  & 79\% & 9\% & 12\% & 1.68 \\
       Not using the goal & 52\% & 16\% & 32\% & 1.20 \\
       \hline
    \end{tabular}
    \label{tab:rel}
    \caption{
     How often the hypotheses proposed by \texttt{text-davinci-003} are rated by the authors as \circled{2}/\circled{1}/\circled{0} in terms of relevance (Section \ref{sec:metrics}).
     Overall, using the goal significantly increases relevance.
    }
\end{table*}

\noindent \textbf{Investigating whether using goals improves relevance on \datasetname{}.}
We then investigated whether \texttt{text-davinci-003} can leverage the goals to propose more relevant hypotheses on more realistic applications in \datasetname{}. 
We sampled 100 problems from \datasetname\phantom{0}with distinct goals and randomly sampled 2 hypotheses from \texttt{text-davinci-003} with/without using goals (see Figure \ref{fig:prompts}), resulting in 400 hypotheses to evaluate.
Three authors then rated their relevance based on the rubric in Section \ref{sec:metrics}, while being blinded about which hypotheses were generated with the goal.
Our main paper focuses on presenting the evaluations performed by ourselves, since crowdworkers might be noisy and untrustworthy \citep{veselovsky2023artificial, suhr2021crowdsourcing}.

We report the results in Table \ref{tab:rel}.
Since this evaluation is subjective, the inter-annotator agreement is only moderate (Kappa=$0.56$);
however, we can still robustly conclude that \texttt{text-davinci-003} can leverage goals to propose hypotheses with higher average relevance rating, since this conclusion can be independently reproduced by every individual evaluator with $p < 10^{-8}$.
To make sure that the same conclusion can be robustly reproduced by external non-authors, we also evaluated the relevance of the hypotheses with Amazon Mechanical Turks, \texttt{gpt-3.5-turbo}, \texttt{Claude-v1.3}, and \texttt{gpt-4}.
We report the results in Appendix Table \ref{tab:external-relevance} and found that our conclusion robustly holds under five different types of evaluators, including expert authors, external crowdworkers, and language models from different companies with different levels of capabilities.

Finally, we conducted similar experiments for the novelty and significance metrics in Appendix \ref{app:more-metrics} and found that they both benefit from using goals as well.
In the next section, we present example discoveries on \datasetname{} to qualitatively understand what a \taskname{} system can achieve. 

\section{Qualitatively Analyzing Discoveries and Limitations with \datasetname{}}

To understand the utility and the limitation of a \taskname{} system, we ran it on \datasetname{}, a set of realistic \taskname{} problems, and analyze the output discoveries qualitatively.

\begin{table}[b!]
    \centering

\begin{tabular}{lcccc}
\toprule
                                         Output discovery &     $V$ &         $p$ &    $V'$ &         $p'$ \\
\midrule
 ``\textit{argues for a path forward to promote the fair ...}'' &  0.16 &  1.26e-04 &  0.35 &   2.01e-73 \\
         ``\textit{refers to illegal immigrants as criminals}'' &  0.09 &  6.17e-03 &  0.19 &   3.17e-38 \\
 ``\textit{has an informal tone, such as slang or colloqu...}'' &  0.08 &  2.35e-03 &  0.24 &   1.46e-35 \\
                          ``\textit{mentions lack of legroom }''&  0.16 &  1.15e-03 &  0.38 &   1.34e-45 \\
                       ``\textit{mentions children or family}'' &  0.08 &  1.00e-05 &  0.11 &   8.05e-09 \\

\bottomrule
\end{tabular}
    \caption{A subset of discoveries presented in Section \ref{sec:application} and their associated estimated validity score $V$, validity score approximated by a model $V'$, and their respective $p$-values $p$ ($p'$) for the null hypothesis that $V(V') < 0$ under a t-test. 
    We present the full set of 13 discoveries in Table \ref{tab:full-discovery-values}.
    }
    \label{tab:partial-discovery-values}
\end{table}

\subsection{Producing Discoveries on \datasetname{} and Analyzing Them} \label{sec:application} 
We ran our \taskname{} system on \datasetname{}, producing 3296 discoveries in total. 
However, we do not have enough budget to validate every finding, since estimating $V$ is expensive (Section \ref{sec:metrics}).
Therefore, from the remaining 3296 discoveries, we manually selected 21 discoveries that 1) achieve a relevance score of \circled{2}, 2) are representative of potential use cases, 3) do not require expert knowledge for Turkers to judge, and 4) are likely to achieve a small $p$-value with fewer than 200 samples from $\mathcal{D}^{\text{val}}$ .

We then estimated their validity based on the procedure described in Section \ref{sec:metrics} by using fewer than 200 samples
%\footnote{We determined the number of samples s.t. $V$ can achieve a $p$-value of $0.005$.} 
from the validation split and calculated the $p$-values, which cost us $\sim$\$1500 in total on MTurk. 
Since we are testing multiple discoveries and each of them can be statistically significant merely due to chance, we keep 13 discoveries with $V$ that are significantly non-zero with $p$-value below 7\%, a threshold determined by the Benjamini Hochberg's procedure with a false discovery rate of 10\%.
In other words, <10\% of the discoveries presented are false discoveries in expectation.

We detail 5 of the 13 discoveries in this section, with the remainder in Appendix \ref{app:more}. 
For each discovery, we report its automated validity score $V'$, the estimated true validity score $V$, and their respective $p$ values in Table \ref{tab:partial-discovery-values}.

\noindent\textbf{Understanding political stances and stereotypes in speeches.}
When comparing presidential speeches on immigrants from Obama to those from Trump, the former  ``\textit{argues for a path forward to promote the fair and just treatment of immigrants}'', while the latter more frequently ``\textit{refers to illegal immigrants as criminals}''.

\noindent\textbf{Analyzing errors in NLP systems.} We fine-tuned a pair of models on two different natural language inference datasets, (a) MNLI and (b) SNLI. 
To understand their patterns of errors, we defined Corpus A to be the subset of MNLI where a is right and b is wrong, and Corpus B to be where b is right and a is wrong.
We found that the latter more often ``\textit{has an informal tone, such as slang or colloquial speech}''.
One possible explanation is that MNLI contains more different genres and hence more informal speeches, causing the former model to perform better on these examples.

\noindent\textbf{Analyzing airline customer reviews.}
We compared the concerns in reviews of the airline Air Canada v.s.~its subsidiary, Air Canada Rogue, which is considered a low-price wing of Air Canada. 
The latter more often ``\textit{mentions lack of legroom}''.

\noindent\textbf{Analyzing gender differences in self-reported happy moments.}
Compared to self-reported happy moments written by males, those by females ``\textit{mentions children or family}'' more often.
{\color{red} Caution:} misinterpreting this correlation as causation could reinforce societal biases (Section \ref{sec:analysis}). 

Due to space constraints, we list more examples on analyzing distribution shifts, text clusters, lyric styles, and news headlines in Appendix \ref{app:more} and their associated $V$ and $V'$ values in Appendix Table \ref{tab:full-discovery-values}.
Across these discoveries, the approximated validity score $V'$ has a 71\% spearman rank correlation with human rating $V$ (66\% for Pearson correlation), thus providing informative yet unreliable signals to practitioners about their validity. 
We hope that $V'$ can better approximate $V$ values in the future as the quality of the validators improve. 
Finally, future works can collect more open problems, allowing \taskname{} systems to produce more impactful discoveries. 
%every time we build a better \taskname{} system in the future, we may use it to automatically generate useful discoveries on an existing aggregation of open problems like \datasetname{} and send the discoveries to researchers who posed the problems.

\subsection{Concrete Examples in \datasetname{} Inform Limitations of \taskname{} Evaluation} \label{sec:analysis}

We discuss limitations of \taskname{} evaluation in this section using concrete examples from \datasetname{}.

\noindent\textbf{Our metrics do not evaluate diversity.}
There are often multiple valid and relevant discoveries, and our system ideally should generate all of them.
For example, when comparing low-rating and high-rating reviews to understand what stands out to customers, both ``\textit{mentions the hidden fees and poor customer service at the airport}'' and ``\textit{mentions the airline charging extra for carry-on items}'' could be valid discoveries.
Our current evaluation does not reward diverse discoveries, and the current system sometimes repeats a discovery using similar paraphrases, e.g., ``\textit{mentions the rude and unprofessional attitude of the staff}'' and ``\textit{mentions the staff being rude and unhelpful}''.
Future evaluation metrics can take diversity into account.

\noindent\textbf{Interpreting discoveries requires domain experts.}
We used Turkers' judgment when computing $T(h, x)$ to judge the validity of a discovery.
However, many discoveries require expert knowledge to interpret properly.
For example, it requires medical training to reliably judge whether a self-reported drug-use experience satisfies ``\textit{mentions psychedelics, such as LSD and shrooms.}''

\noindent\textbf{Correlation $\neq$ causation.}
Our metrics currently do not evaluate whether the discovery is causally related to how the corpus pair was generated.
For example, when comparing self-reported happy moments from females and males, even if the former corpus has more samples that ``\textit{mention children and family}'', it does not necessarily imply family plays a more important role in inter-personal relations for females; an alternative hypothesis is that females might mention people in general more often than males do, hence leading to the observation that they mention family more often. 
Spurious correlations could also sneak into our validity evaluation: for example, if the Turkers implicitly associate female activities as family-related \cite{greenwald1995implicit}, then we might falsely make this discovery due to evaluator biases.
Future metrics should also consider plausible alternative hypotheses to evaluate causality and control the potential biases from the human evaluators.
Additionally, we should treat the discovery from \taskname{} with caution to prevent automating and amplifying societal biases. 

We discuss other limitations, such as restricting the discovery to be a single predicate, the biases in authors' qualitative evaluation, and the incomprehensiveness of \datasetname{} in Appendix \ref{app:limit}.

\section{Related Work and Discussion} \label{sec:related}

\noindent\textbf{Inductive Reasoning with NLP Models.}
Recent works show that language models are capable of inductive reasoning under restricted settings, discovering patterns from a set of text data points and describing them with language \citep{ honovich2022instruction}.
\citet{yang2022language} use this capability to induce natural language rules with the format of ``\textit{if \dots then \dots}''.
\citet{zhou2022large} and \citet{ye2022guess} use this capability to improve zero/few-shot accuracy by inferring the most likely instruction using input-output example(s) of the target task.
\citet{zhong2022describing} and \citet{singh2022explaining} use this capability to discover patterns in datasets, and we improve by building an automatic benchmark and a dataset of open-ended problems and require the discovery to be relevant.

ML models can also perform inductive reasoning in other modalities, such as vision.
\citet{hernandez2021natural} describes visual features that activate a neuron; 
\citet{zhu2022gsclip} describes distribution shifts between the training distribution and the test distribution for images; 
and \citet{eyuboglu2022domino} describes errors made by vision models.
We hope future models can perform inductive reasoning in other modalities, such as sound \citep{aghajanyan2023scaling} or physical senses \citep{thomason2016learning}.

\noindent\textbf{Exploratory Analysis and Automated Discovery.}
It is not new to automatically discover patterns by learning from empirical data.
To list a few classical methods, linear regression analyzes the effect of each real-valued feature by interpreting the learned weights \citep{draper1998applied};
n-gram models can extract discriminative phrases, thus yielding insights about corpus-level differences \citep{manning1999foundations};
topic models \citep{blei2003latent} can extract major topical variations across documents, where each topic is represented as a distribution over words;
small decision trees can extract interpretable if-then statements \citep{2015};
and an entity embedding model learned on existing relations between entities can predict unseen relations \citep{socher2013reasoning}. 
In comparison, \taskname{} produces discoveries in the form of natural language predicates, which are interpretable and can express abstract concepts; additionally, it is more directed at the goal, while machine learning classifiers like naïve bayes or linear regression will pick up any discriminative features: Appendix \ref{app:bayes} offers a more comprehensive discussion using examples from \syndataset{}. 
Given the respective strength of \taskname{} and traditional exploratory methods, we envision \taskname{} to serve as a complementary method to traditional methods.

\noindent\textbf{Epistemology.} While the process of validating a hypothesis is well-formulated, it is much less well-understood how to automatically generate hypotheses and decide what discoveries are meaningful \citep{shapere1964structure, heckman2017abducting}.
Related works in this area have been sparse, among which \citet{mcgarry2005survey} sketches high-level principles for evaluating knowledge discoveries and \citet{ludwig2022algorithmic} proposes to crowd-source hypotheses from MTurk workers.
We concur with the perspective of \citet{polanyi2000republic} that meaningfulness of a hypothesis cannot be explicitly verbalized with simple logic but is dependent on implicit community norms;
therefore, the process of proposing hypotheses should be learned from empirical data (e.g. pre-training, self-training, or human feedback) rather than deduced from a priori analysis of concepts \citep{quine1969naturalistic}. 
We hope contributions from other domains can provide more empirical data on what discoveries are meaningful, hence guiding our system to produce more important discoveries.

\section*{Acknowledgement}
We thank Xiaochuang Han and Sam Bowman for their early discussions on this project.
We thank Cathy Chen, Erik Jones, Jessy Lin, Alex Pan, Chenglei Si, Xi Ye, and Tianyi Zhang for their helpful feedback on the paper draft.
We thank OpenAI and Anthropic for providing model access.

\bibliographystyle{iclr2023_conference}
\bibliography{example_paper}

\begin{thebibliography}{91}
\providecommand{\natexlab}[1]{#1}
\providecommand{\url}[1]{\texttt{#1}}
\expandafter\ifx\csname urlstyle\endcsname\relax
  \providecommand{\doi}[1]{doi: #1}\else
  \providecommand{\doi}{doi: \begingroup \urlstyle{rm}\Url}\fi

\bibitem[ess(2012)]{essay-scoring}
The hewlett foundation: Automated essay scoring, 2012.
\newblock URL \url{https://kaggle.com/competitions/asap-aes}.

\bibitem[sho(2013)]{short-answer-scoring}
The hewlett foundation: Short answer scoring, 2013.
\newblock URL \url{https://kaggle.com/competitions/asap-sas}.

\bibitem[pol(2021)]{political-ads}
Ad observer.
\newblock https://adobserver.org/, 2021.
\newblock Accessed: 2022-12-30.

\bibitem[Aghajanyan et~al.(2023)Aghajanyan, Yu, Conneau, Hsu, Hambardzumyan,
  Zhang, Roller, Goyal, Levy, and Zettlemoyer]{aghajanyan2023scaling}
Armen Aghajanyan, Lili Yu, Alexis Conneau, Wei-Ning Hsu, Karen Hambardzumyan,
  Susan Zhang, Stephen Roller, Naman Goyal, Omer Levy, and Luke Zettlemoyer.
\newblock Scaling laws for generative mixed-modal language models.
\newblock \emph{arXiv preprint arXiv:2301.03728}, 2023.

\bibitem[Aharoni \& Goldberg(2020)Aharoni and
  Goldberg]{aharoni-goldberg-2020-unsupervised}
Roee Aharoni and Yoav Goldberg.
\newblock Unsupervised domain clusters in pretrained language models.
\newblock In \emph{Proceedings of the 58th Annual Meeting of the Association
  for Computational Linguistics}, pp.\  7747--7763, Online, July 2020.
  Association for Computational Linguistics.
\newblock \doi{10.18653/v1/2020.acl-main.692}.
\newblock URL \url{https://aclanthology.org/2020.acl-main.692}.

\bibitem[Alali et~al.(2021)Alali, Syed, Alsayed, Patel, and
  Bodala]{scotus-cases}
Mohammad Alali, Shaayan Syed, Mohammed Alsayed, Smit Patel, and Hemanth Bodala.
\newblock Justice: A benchmark dataset for supreme court's judgment prediction.
\newblock \emph{arXiv preprint arXiv:2112.03414}, 2021.

\bibitem[Asai et~al.(2018)Asai, Evensen, Golshan, Halevy, Li, Lopatenko,
  Stepanov, Suhara, Tan, and Xu]{happy-moments}
Akari Asai, Sara Evensen, Behzad Golshan, Alon Halevy, Vivian Li, Andrei
  Lopatenko, Daniela Stepanov, Yoshihiko Suhara, Wang-Chiew Tan, and Yinzhan
  Xu.
\newblock Happydb: A corpus of 100,000 crowdsourced happy moments.
\newblock \emph{arXiv preprint arXiv:1801.07746}, 2018.

\bibitem[Bai et~al.(2022{\natexlab{a}})Bai, Jones, Ndousse, Askell, Chen,
  DasSarma, Drain, Fort, Ganguli, Henighan, et~al.]{bai2022training}
Yuntao Bai, Andy Jones, Kamal Ndousse, Amanda Askell, Anna Chen, Nova DasSarma,
  Dawn Drain, Stanislav Fort, Deep Ganguli, Tom Henighan, et~al.
\newblock Training a helpful and harmless assistant with reinforcement learning
  from human feedback.
\newblock \emph{arXiv preprint arXiv:2204.05862}, 2022{\natexlab{a}}.

\bibitem[Bai et~al.(2022{\natexlab{b}})Bai, Kadavath, Kundu, Askell, Kernion,
  Jones, Chen, Goldie, Mirhoseini, McKinnon, et~al.]{bai2022constitutional}
Yuntao Bai, Saurav Kadavath, Sandipan Kundu, Amanda Askell, Jackson Kernion,
  Andy Jones, Anna Chen, Anna Goldie, Azalia Mirhoseini, Cameron McKinnon,
  et~al.
\newblock Constitutional ai: Harmlessness from ai feedback.
\newblock \emph{arXiv preprint arXiv:2212.08073}, 2022{\natexlab{b}}.

\bibitem[Barnum \& Lo(2020)Barnum and Lo]{npt-conferences}
Miriam Barnum and James Lo.
\newblock Is the npt unraveling? evidence from text analysis of review
  conference statements.
\newblock \emph{Journal of Peace Research}, 57\penalty0 (6):\penalty0 740--751,
  2020.

\bibitem[Baturo et~al.(2017)Baturo, Dasandi, and Mikhaylov]{un-debates}
Alexander Baturo, Niheer Dasandi, and Slava~J Mikhaylov.
\newblock Understanding state preferences with text as data: Introducing the un
  general debate corpus.
\newblock \emph{Research \& Politics}, 4\penalty0 (2):\penalty0
  2053168017712821, 2017.

\bibitem[Bhalotia(2022)]{yc-startups}
Akshay Bhalotia.
\newblock Yc company scraper.
\newblock \url{https://github.com/akshaybhalotia/yc_company_scraper}, 2022.

\bibitem[Bird et~al.(2009)Bird, Klein, and Loper]{bird2009natural}
Steven Bird, Ewan Klein, and Edward Loper.
\newblock \emph{Natural language processing with Python: analyzing text with
  the natural language toolkit}.
\newblock " O'Reilly Media, Inc.", 2009.

\bibitem[Blei et~al.(2003)Blei, Ng, and Jordan]{blei2003latent}
David~M Blei, Andrew~Y Ng, and Michael~I Jordan.
\newblock Latent dirichlet allocation.
\newblock \emph{Journal of machine Learning research}, 3\penalty0
  (Jan):\penalty0 993--1022, 2003.

\bibitem[Bowman et~al.(2015)Bowman, Angeli, Potts, and Manning]{snli}
Samuel~R Bowman, Gabor Angeli, Christopher Potts, and Christopher~D Manning.
\newblock The snli corpus.
\newblock 2015.

\bibitem[Bramhecha(2019)]{poetry}
Divy Bramhecha.
\newblock {Poetry Foundation Poems}, 2019.
\newblock URL
  \url{https://www.kaggle.com/datasets/tgdivy/poetry-foundation-poems}.

\bibitem[Card et~al.(2022)Card, Chang, Becker, Mendelsohn, Voigt, Boustan,
  Abramitzky, and Jurafsky]{immigration-speeches}
Dallas Card, Serina Chang, Chris Becker, Julia Mendelsohn, Rob Voigt, Leah
  Boustan, Ran Abramitzky, and Dan Jurafsky.
\newblock Replication code and data for ``{C}omputational analysis of 140 years
  of {US} political speeches reveals more positive but increasingly polarized
  framing of immigration'' [dataset].
\newblock \emph{https://github.com/dallascard/us-immigration-speeches/}, 2022.

\bibitem[Chalkidis et~al.(2019)Chalkidis, Androutsopoulos, and
  Aletras]{echr-decisions}
Ilias Chalkidis, Ion Androutsopoulos, and Nikolaos Aletras.
\newblock Neural legal judgment prediction in english.
\newblock \emph{arXiv preprint arXiv:1906.02059}, 2019.

\bibitem[Chen et~al.(2019)Chen, Khashabi, Yin, Callison-Burch, and
  Roth]{CKYCR19}
Sihao Chen, Daniel Khashabi, Wenpeng Yin, Chris Callison-Burch, and Dan Roth.
\newblock {Seeing Things from a Different Angle: Discovering Diverse
  Perspectives about Claims}.
\newblock In \emph{Proc. of the Annual Conference of the North American Chapter
  of the Association for Computational Linguistics (NAACL)}, 2019.
\newblock URL \url{http://cogcomp.org/papers/CKYCR19.pdf}.

\bibitem[Chung et~al.(2022)Chung, Hou, Longpre, Zoph, Tay, Fedus, Li, Wang,
  Dehghani, Brahma, et~al.]{chung2022scaling}
Hyung~Won Chung, Le~Hou, Shayne Longpre, Barret Zoph, Yi~Tay, William Fedus,
  Eric Li, Xuezhi Wang, Mostafa Dehghani, Siddhartha Brahma, et~al.
\newblock Scaling instruction-finetuned language models.
\newblock \emph{arXiv preprint arXiv:2210.11416}, 2022.

\bibitem[Clark et~al.(2019)Clark, Lee, Chang, Kwiatkowski, Collins, and
  Toutanova]{boolq}
Christopher Clark, Kenton Lee, Ming-Wei Chang, Tom Kwiatkowski, Michael
  Collins, and Kristina Toutanova.
\newblock Boolq: Exploring the surprising difficulty of natural yes/no
  questions.
\newblock \emph{arXiv preprint arXiv:1905.10044}, 2019.

\bibitem[Draper \& Smith(1998)Draper and Smith]{draper1998applied}
Norman~R Draper and Harry Smith.
\newblock \emph{Applied regression analysis}, volume 326.
\newblock John Wiley \& Sons, 1998.

\bibitem[Eyuboglu et~al.(2022)Eyuboglu, Varma, Saab, Delbrouck, Lee-Messer,
  Dunnmon, Zou, and R{\'e}]{eyuboglu2022domino}
Sabri Eyuboglu, Maya Varma, Khaled Saab, Jean-Benoit Delbrouck, Christopher
  Lee-Messer, Jared Dunnmon, James Zou, and Christopher R{\'e}.
\newblock Domino: Discovering systematic errors with cross-modal embeddings.
\newblock \emph{arXiv preprint arXiv:2203.14960}, 2022.

\bibitem[Gao et~al.(2021)Gao, Jang, and Yang]{parenting-subreddits}
Yujia Gao, Jinu Jang, and Diyi Yang.
\newblock Understanding the usage of online media for parenting from infancy to
  preschool at scale.
\newblock In \emph{Proceedings of the 2021 CHI Conference on Human Factors in
  Computing Systems}, pp.\  1--12, 2021.

\bibitem[Greenwald \& Banaji(1995)Greenwald and Banaji]{greenwald1995implicit}
Anthony~G Greenwald and Mahzarin~R Banaji.
\newblock Implicit social cognition: attitudes, self-esteem, and stereotypes.
\newblock \emph{Psychological review}, 102\penalty0 (1):\penalty0 4, 1995.

\bibitem[Habernal \& Gurevych(2016)Habernal and Gurevych]{convincing-arguments}
Ivan Habernal and Iryna Gurevych.
\newblock {Which argument is more convincing? Analyzing and predicting
  convincingness of Web arguments using bidirectional LSTM}.
\newblock In \emph{Proceedings of the 54th Annual Meeting of the Association
  for Computational Linguistics (Volume 1: Long Papers)}, pp.\  1589--1599,
  Berlin, Germany, 2016. Association for Computational Linguistics.
\newblock URL \url{http://www.aclweb.org/anthology/P16-1150}.

\bibitem[Hartman(2019)]{ad-transcripts}
Kevin Hartman.
\newblock {Advertisement Transcripts from Various Industries}, 2019.
\newblock URL \url{https://tinyurl.com/5w36dwdx}.

\bibitem[He et~al.(2018)He, Chen, Balakrishnan, and
  Liang]{craigslist-negotiations}
He~He, Derek Chen, Anusha Balakrishnan, and Percy Liang.
\newblock Decoupling strategy and generation in negotiation dialogues, 2018.

\bibitem[He(2020)]{rate-my-prof}
Jibo He.
\newblock {Big Data Set from RateMyProfessor.com for Professors' Teaching
  Evaluation}, 2020.
\newblock URL \url{https://data.mendeley.com/datasets/fvtfjyvw7d/2}.

\bibitem[He(2021)]{suicide-notes}
Samuel He.
\newblock Goodbye world: using natural language processing to identify suicidal
  posts, 2021.
\newblock URL \url{https://github.com/hesamuel/goodbye_world}.

\bibitem[Heckman \& Singer(2017)Heckman and Singer]{heckman2017abducting}
James~J Heckman and Burton Singer.
\newblock Abducting economics.
\newblock \emph{American Economic Review}, 107\penalty0 (5):\penalty0 298--302,
  2017.

\bibitem[Hernandez et~al.(2021)Hernandez, Schwettmann, Bau, Bagashvili,
  Torralba, and Andreas]{hernandez2021natural}
Evan Hernandez, Sarah Schwettmann, David Bau, Teona Bagashvili, Antonio
  Torralba, and Jacob Andreas.
\newblock Natural language descriptions of deep visual features.
\newblock In \emph{International Conference on Learning Representations}, 2021.

\bibitem[Honovich et~al.(2022)Honovich, Shaham, Bowman, and
  Levy]{honovich2022instruction}
Or~Honovich, Uri Shaham, Samuel~R Bowman, and Omer Levy.
\newblock Instruction induction: From few examples to natural language task
  descriptions.
\newblock \emph{arXiv preprint arXiv:2205.10782}, 2022.

\bibitem[Hossain et~al.(2019)Hossain, Krumm, and Gamon]{microedit-humor}
Nabil Hossain, John Krumm, and Michael Gamon.
\newblock " president vows to cut< taxes> hair": Dataset and analysis of
  creative text editing for humorous headlines.
\newblock \emph{arXiv preprint arXiv:1906.00274}, 2019.

\bibitem[Kaggle(2018)]{movie-tmdb}
Kaggle.
\newblock {TMDB 5000 Movie Dataset}, 2018.
\newblock URL \url{https://www.kaggle.com/datasets/tmdb/tmdb-movie-metadata}.

\bibitem[Kulkarni(2018)]{abc-headlines}
Rohit Kulkarni.
\newblock {A Million News Headlines}, 2018.
\newblock URL \url{https://doi.org/10.7910/DVN/SYBGZL}.

\bibitem[Kulkarni(2020{\natexlab{a}})]{clickbait-headlines}
Rohit Kulkarni.
\newblock {The Examiner - Spam Clickbait Catalog}, 2020{\natexlab{a}}.
\newblock URL
  \url{https://www.kaggle.com/datasets/therohk/examine-the-examiner}.

\bibitem[Kulkarni(2020{\natexlab{b}})]{urban-dictionary}
Rohit Kulkarni.
\newblock {Urban Dictionary Words And Definitions}, 2020{\natexlab{b}}.
\newblock URL
  \url{https://www.kaggle.com/datasets/therohk/urban-dictionary-words-dataset}.

\bibitem[Kulkarni(2022)]{times-india-headlines}
Rohit Kulkarni.
\newblock {India News Headlines Dataset}, 2022.
\newblock URL
  \url{https://www.kaggle.com/datasets/therohk/india-headlines-news-dataset}.

\bibitem[Letham et~al.(2015)Letham, Rudin, McCormick, and Madigan]{2015}
Benjamin Letham, Cynthia Rudin, Tyler~H. McCormick, and David Madigan.
\newblock Interpretable classifiers using rules and bayesian analysis: Building
  a better stroke prediction model.
\newblock \emph{The Annals of Applied Statistics}, 9\penalty0 (3), Sep 2015.
\newblock ISSN 1932-6157.
\newblock \doi{10.1214/15-aoas848}.
\newblock URL \url{http://dx.doi.org/10.1214/15-AOAS848}.

\bibitem[Lim \& Benson(2021)Lim and Benson]{genius-lyrics}
Derek Lim and Austin~R Benson.
\newblock Expertise and dynamics within crowdsourced musical knowledge
  curation: A case study of the genius platform.
\newblock In \emph{ICWSM}, pp.\  373--384, 2021.

\bibitem[Liu et~al.(2022)Liu, Swayamdipta, Smith, and Choi]{nli-benchmarks}
Alisa Liu, Swabha Swayamdipta, Noah~A. Smith, and Yejin Choi.
\newblock Wanli: Worker and ai collaboration for natural language inference
  dataset creation, January 2022.
\newblock URL \url{https://arxiv.org/pdf/2201.05955}.

\bibitem[Liu et~al.(2019)Liu, Ott, Goyal, Du, Joshi, Chen, Levy, Lewis,
  Zettlemoyer, and Stoyanov]{liu2019roberta}
Yinhan Liu, Myle Ott, Naman Goyal, Jingfei Du, Mandar Joshi, Danqi Chen, Omer
  Levy, Mike Lewis, Luke Zettlemoyer, and Veselin Stoyanov.
\newblock Roberta: A robustly optimized bert pretraining approach.
\newblock \emph{arXiv preprint arXiv:1907.11692}, 2019.

\bibitem[Liu(2011)]{reuters-authorship}
Zhi Liu.
\newblock {Reuter\_50\_50 Data Set}, 2011.
\newblock URL \url{https://archive.ics.uci.edu/ml/datasets/Reuter_50_50}.

\bibitem[Ludwig \& Mullainathan(2022)Ludwig and
  Mullainathan]{ludwig2022algorithmic}
Jens Ludwig and Sendhil Mullainathan.
\newblock Algorithmic behavioral science: Machine learning as a tool for
  scientific discovery.
\newblock \emph{Chicago Booth Research Paper}, \penalty0 (22-15), 2022.

\bibitem[Manning \& Schutze(1999)Manning and Schutze]{manning1999foundations}
Christopher Manning and Hinrich Schutze.
\newblock \emph{Foundations of statistical natural language processing}.
\newblock MIT press, 1999.

\bibitem[McGarry(2005)]{mcgarry2005survey}
Ken McGarry.
\newblock A survey of interestingness measures for knowledge discovery.
\newblock \emph{The knowledge engineering review}, 20\penalty0 (1):\penalty0
  39--61, 2005.

\bibitem[Merity et~al.(2016)Merity, Xiong, Bradbury, and Socher]{wikitext}
Stephen Merity, Caiming Xiong, James Bradbury, and Richard Socher.
\newblock Pointer sentinel mixture models, 2016.

\bibitem[Mish(2020)]{fomc-speeches}
Natan Mish.
\newblock {Federal Reserve Governors Speeches 1996 - 2020}, 2020.
\newblock URL \url{https://tinyurl.com/3j2e79a6}.

\bibitem[Mishra et~al.(2022)Mishra, Khashabi, Baral, and
  Hajishirzi]{ai2-natural-instruction}
Swaroop Mishra, Daniel Khashabi, Chitta Baral, and Hannaneh Hajishirzi.
\newblock Cross-task generalization via natural language crowdsourcing
  instructions.
\newblock In \emph{ACL}, 2022.

\bibitem[Misra \& Arora(2019)Misra and Arora]{huff-post-headlines}
Rishabh Misra and Prahal Arora.
\newblock Sarcasm detection using hybrid neural network.
\newblock \emph{arXiv preprint arXiv:1908.07414}, 2019.

\bibitem[Misra \& Grover(2021)Misra and Grover]{huff-post-headlines2}
Rishabh Misra and Jigyasa Grover.
\newblock \emph{Sculpting Data for ML: The first act of Machine Learning}.
\newblock 01 2021.
\newblock ISBN 9798585463570.

\bibitem[Moniz \& Torgo(2018)Moniz and Torgo]{news-popularity}
Nuno Moniz and Luâ€™is Torgo.
\newblock Multi-source social feedback of online news feeds.
\newblock \emph{CoRR}, [Web Link], 2018.

\bibitem[Mouillé(2017)]{kickstarter}
Mickaël Mouillé.
\newblock {Kickstarter Projects}, 2017.
\newblock URL
  \url{https://www.kaggle.com/datasets/kemical/kickstarter-projects?select=ks-projects-201612.csv}.

\bibitem[Nguyen et~al.(2020)Nguyen, Liakata, DeDeo, Eisenstein, Mimno, Tromble,
  and Winters]{nguyen2020we}
Dong Nguyen, Maria Liakata, Simon DeDeo, Jacob Eisenstein, David Mimno, Rebekah
  Tromble, and Jane Winters.
\newblock How we do things with words: Analyzing text as social and cultural
  data.
\newblock \emph{Frontiers in Artificial Intelligence}, 3:\penalty0 62, 2020.

\bibitem[Ni et~al.(2019)Ni, Li, and McAuley]{amazon-reviews}
Jianmo Ni, Jiacheng Li, and Julian McAuley.
\newblock Justifying recommendations using distantly-labeled reviews and
  fine-grained aspects.
\newblock In \emph{Proceedings of the 2019 conference on empirical methods in
  natural language processing and the 9th international joint conference on
  natural language processing (EMNLP-IJCNLP)}, pp.\  188--197, 2019.

\bibitem[O'Brien(2020)]{aita}
Elle O'Brien.
\newblock {iterative/aita\_dataset: Praw rescrape of entire dataset}, February
  2020.
\newblock URL \url{https://doi.org/10.5281/zenodo.3677563}.

\bibitem[Ouyang et~al.(2022)Ouyang, Wu, Jiang, Almeida, Wainwright, Mishkin,
  Zhang, Agarwal, Slama, Ray, et~al.]{ouyang2022training}
Long Ouyang, Jeff Wu, Xu~Jiang, Diogo Almeida, Carroll~L Wainwright, Pamela
  Mishkin, Chong Zhang, Sandhini Agarwal, Katarina Slama, Alex Ray, et~al.
\newblock Training language models to follow instructions with human feedback.
\newblock \emph{arXiv preprint arXiv:2203.02155}, 2022.

\bibitem[P{\'e}rez-Rosas \& Mihalcea(2015)P{\'e}rez-Rosas and
  Mihalcea]{open-deception}
Ver{\'o}nica P{\'e}rez-Rosas and Rada Mihalcea.
\newblock Experiments in open domain deception detection.
\newblock In \emph{Proceedings of the 2015 conference on empirical methods in
  natural language processing}, pp.\  1120--1125, 2015.

\bibitem[P{\'e}rez-Rosas et~al.(2015)P{\'e}rez-Rosas, Abouelenien, Mihalcea,
  and Burzo]{trial-deception}
Ver{\'o}nica P{\'e}rez-Rosas, Mohamed Abouelenien, Rada Mihalcea, and Mihai
  Burzo.
\newblock Deception detection using real-life trial data.
\newblock In \emph{Proceedings of the 2015 ACM on International Conference on
  Multimodal Interaction}, pp.\  59--66, 2015.

\bibitem[P{\'e}rez-Rosas et~al.(2017)P{\'e}rez-Rosas, Kleinberg, Lefevre, and
  Mihalcea]{fake-news}
Ver{\'o}nica P{\'e}rez-Rosas, Bennett Kleinberg, Alexandra Lefevre, and Rada
  Mihalcea.
\newblock Automatic detection of fake news.
\newblock \emph{arXiv preprint arXiv:1708.07104}, 2017.

\bibitem[Peskov et~al.(2020)Peskov, Cheng, Elgohary, Barrow,
  Danescu-Niculescu-Mizil, and Boyd-Graber]{diplomacy-deception}
Denis Peskov, Benny Cheng, Ahmed Elgohary, Joe Barrow, Cristian
  Danescu-Niculescu-Mizil, and Jordan Boyd-Graber.
\newblock It takes two to lie: One to lie and one to listen.
\newblock In \emph{Association for Computational Linguistics}, 2020.

\bibitem[Pestian et~al.(2007)Pestian, Brew, Matykiewicz, Hovermale, Johnson,
  Cohen, and Duch]{radiology-diagnosis}
John~P. Pestian, Chris Brew, Pawel Matykiewicz, DJ~Hovermale, Neil Johnson,
  K.~Bretonnel Cohen, and Wlodzislaw Duch.
\newblock A shared task involving multi-label classification of clinical free
  text.
\newblock In \emph{Biological, translational, and clinical language
  processing}, pp.\  97--104, Prague, Czech Republic, June 2007. Association
  for Computational Linguistics.
\newblock URL \url{https://aclanthology.org/W07-1013}.

\bibitem[Polanyi et~al.(2000)Polanyi, Ziman, and Fuller]{polanyi2000republic}
Michael Polanyi, John Ziman, and Steve Fuller.
\newblock The republic of science: its political and economic theory minerva, i
  (1)(1962), 54-73.
\newblock \emph{Minerva}, 38\penalty0 (1):\penalty0 1--32, 2000.

\bibitem[Price et~al.(2020)Price, Gifford-Moore, Fleming, Musker, Roichman,
  Sylvain, Thain, Dixon, and Sorensen]{unhealthy-conversations}
Ilan Price, Jordan Gifford-Moore, Jory Fleming, Saul Musker, Maayan Roichman,
  Guillaume Sylvain, Nithum Thain, Lucas Dixon, and Jeffrey Sorensen.
\newblock Six attributes of unhealthy conversation.
\newblock \emph{arXiv preprint arXiv:2010.07410}, 2020.

\bibitem[Progress(2022)]{admin-statements}
Demand Progress.
\newblock {Statements of Administration Policy}, 2022.
\newblock URL
  \url{https://github.com/unitedstates/statements-of-administration-policy#statements-of-administration-policy}.

\bibitem[PromptCloud(2017)]{dice-jobs}
PromptCloud.
\newblock {U.S. Technology Jobs on Dice.com}, 2017.
\newblock URL
  \url{https://www.kaggle.com/datasets/PromptCloudHQ/us-technology-jobs-on-dicecom}.

\bibitem[Quine(1969)]{quine1969naturalistic}
WVO Quine.
\newblock Naturalistic epistemology.
\newblock \emph{Ontological relativity and other essays}, pp.\  69--90, 1969.

\bibitem[Quora(2017)]{qqp}
Quora.
\newblock {Quora Question Pairs}, 2017.
\newblock URL \url{https://www.kaggle.com/c/quora-question-pairs}.

\bibitem[Rajpurkar et~al.(2018)Rajpurkar, Jia, and Liang]{squad-v2}
Pranav Rajpurkar, Robin Jia, and Percy Liang.
\newblock Know what you don't know: Unanswerable questions for squad.
\newblock \emph{arXiv preprint arXiv:1806.03822}, 2018.

\bibitem[Robischon(2019)]{movie-wiki}
Justin Robischon.
\newblock {Wikipedia Movie Plots}, 2019.
\newblock URL
  \url{https://www.kaggle.com/datasets/jrobischon/wikipedia-movie-plots}.

\bibitem[Roush \& Balaji(2020)Roush and Balaji]{debate}
Allen Roush and Arvind Balaji.
\newblock {D}ebate{S}um: A large-scale argument mining and summarization
  dataset.
\newblock In \emph{Proceedings of the 7th Workshop on Argument Mining}, pp.\
  1--7, Online, December 2020. Association for Computational Linguistics.
\newblock URL \url{https://aclanthology.org/2020.argmining-1.1}.

\bibitem[Shapere(1964)]{shapere1964structure}
Dudley Shapere.
\newblock The structure of scientific revolutions.
\newblock \emph{The Philosophical Review}, 73\penalty0 (3):\penalty0 383--394,
  1964.

\bibitem[Singh et~al.(2022)Singh, Morris, Aneja, Rush, and
  Gao]{singh2022explaining}
Chandan Singh, John~X Morris, Jyoti Aneja, Alexander~M Rush, and Jianfeng Gao.
\newblock Explaining patterns in data with language models via interpretable
  autoprompting.
\newblock \emph{arXiv preprint arXiv:2210.01848}, 2022.

\bibitem[Socher et~al.(2013)Socher, Chen, Manning, and Ng]{socher2013reasoning}
Richard Socher, Danqi Chen, Christopher~D Manning, and Andrew Ng.
\newblock Reasoning with neural tensor networks for knowledge base completion.
\newblock \emph{Advances in neural information processing systems}, 26, 2013.

\bibitem[Suhr et~al.(2021)Suhr, Vania, Nangia, Sap, Yatskar, Bowman, and
  Artzi]{suhr2021crowdsourcing}
Alane Suhr, Clara Vania, Nikita Nangia, Maarten Sap, Mark Yatskar, Samuel
  Bowman, and Yoav Artzi.
\newblock Crowdsourcing beyond annotation: Case studies in benchmark data
  collection.
\newblock In \emph{Proceedings of the 2021 Conference on Empirical Methods in
  Natural Language Processing: Tutorial Abstracts}, pp.\  1--6, 2021.

\bibitem[Sun(2017)]{stock-news}
J~Sun.
\newblock {Daily News for Stock Market Prediction}, 2017.
\newblock URL \url{https://www.kaggle.com/datasets/aaron7sun/stocknews}.

\bibitem[Thomason et~al.(2016)Thomason, Sinapov, Svetlik, Stone, and
  Mooney]{thomason2016learning}
Jesse Thomason, Jivko Sinapov, Maxwell Svetlik, Peter Stone, and Raymond~J
  Mooney.
\newblock Learning multi-modal grounded linguistic semantics by playing" i
  spy".
\newblock In \emph{IJCAI}, pp.\  3477--3483, 2016.

\bibitem[Thompson(2019)]{all-the-news}
Andrew Thompson.
\newblock {All the News 1.0}, 2019.
\newblock URL
  \url{https://components.one/datasets/all-the-news-articles-dataset}.

\bibitem[Turcan \& McKeown(2019)Turcan and McKeown]{reddit-stress}
Elsbeth Turcan and Kathleen McKeown.
\newblock Dreaddit: A reddit dataset for stress analysis in social media.
\newblock \emph{arXiv preprint arXiv:1911.00133}, 2019.

\bibitem[Udacity(2017)]{armenian-jobs}
Udacity.
\newblock {Armenian Online Job Postings}, 2017.
\newblock URL
  \url{https://www.kaggle.com/datasets/udacity/armenian-online-job-postings}.

\bibitem[Veselovsky et~al.(2023)Veselovsky, Ribeiro, and
  West]{veselovsky2023artificial}
Veniamin Veselovsky, Manoel~Horta Ribeiro, and Robert West.
\newblock Artificial artificial artificial intelligence: Crowd workers widely
  use large language models for text production tasks.
\newblock \emph{arXiv preprint arXiv:2306.07899}, 2023.

\bibitem[Wang et~al.(2022)Wang, Mishra, Alipoormolabashi, Kordi, Mirzaei,
  Arunkumar, Ashok, Dhanasekaran, Naik, Stap, et~al.]{wang2022super}
Yizhong Wang, Swaroop Mishra, Pegah Alipoormolabashi, Yeganeh Kordi, Amirreza
  Mirzaei, Anjana Arunkumar, Arjun Ashok, Arut~Selvan Dhanasekaran, Atharva
  Naik, David Stap, et~al.
\newblock Super-naturalinstructions: Generalization via declarative
  instructions on 1600+ nlp tasks.
\newblock \emph{URL https://arxiv. org/abs/2204.07705}, 2022.

\bibitem[Wang et~al.(2023)Wang, Shang, and Zhong]{wang2023goal}
Zihan Wang, Jingbo Shang, and Ruiqi Zhong.
\newblock Goal-driven explainable clustering via language descriptions.
\newblock \emph{arXiv preprint arXiv:2305.13749}, 2023.

\bibitem[Weller \& Seppi(2020)Weller and Seppi]{reddit-humor}
Orion Weller and Kevin Seppi.
\newblock The r{J}okes dataset: a large scale humor collection.
\newblock In \emph{Proceedings of the Twelfth Language Resources and Evaluation
  Conference}, pp.\  6136--6141, Marseille, France, May 2020. European Language
  Resources Association.
\newblock ISBN 979-10-95546-34-4.
\newblock URL \url{https://aclanthology.org/2020.lrec-1.753}.

\bibitem[Williams et~al.(2017)Williams, Nangia, and Bowman]{mnli}
Adina Williams, Nikita Nangia, and Samuel~R Bowman.
\newblock A broad-coverage challenge corpus for sentence understanding through
  inference.
\newblock \emph{arXiv preprint arXiv:1704.05426}, 2017.

\bibitem[Yang et~al.(2022)Yang, Dong, Du, Cheng, Cambria, Liu, Gao, and
  Wei]{yang2022language}
Zonglin Yang, Li~Dong, Xinya Du, Hao Cheng, Erik Cambria, Xiaodong Liu,
  Jianfeng Gao, and Furu Wei.
\newblock Language models as inductive reasoners.
\newblock \emph{arXiv preprint arXiv:2212.10923}, 2022.

\bibitem[Ye et~al.(2022)Ye, Kim, Jang, Shin, and Seo]{ye2022guess}
Seonghyeon Ye, Doyoung Kim, Joel Jang, Joongbo Shin, and Minjoon Seo.
\newblock Guess the instruction! making language models stronger zero-shot
  learners.
\newblock \emph{arXiv preprint arXiv:2210.02969}, 2022.

\bibitem[Zhong et~al.(2022)Zhong, Snell, Klein, and
  Steinhardt]{zhong2022describing}
Ruiqi Zhong, Charlie Snell, Dan Klein, and Jacob Steinhardt.
\newblock Describing differences between text distributions with natural
  language.
\newblock In \emph{International Conference on Machine Learning}, pp.\
  27099--27116. PMLR, 2022.

\bibitem[Zhou et~al.(2022)Zhou, Muresanu, Han, Paster, Pitis, Chan, and
  Ba]{zhou2022large}
Yongchao Zhou, Andrei~Ioan Muresanu, Ziwen Han, Keiran Paster, Silviu Pitis,
  Harris Chan, and Jimmy Ba.
\newblock Large language models are human-level prompt engineers.
\newblock \emph{arXiv preprint arXiv:2211.01910}, 2022.

\bibitem[Zhu et~al.(2022)Zhu, Liang, and Zou]{zhu2022gsclip}
Zhiying Zhu, Weixin Liang, and James Zou.
\newblock Gsclip: A framework for explaining distribution shifts in natural
  language.
\newblock \emph{arXiv preprint arXiv:2206.15007}, 2022.

\end{thebibliography}
\pagebreak

\section{Cost for Running the Experiments} \label{app:compute}

For each problem, we ran the proposer for ~10 times on average; assuming each prompt to be at most 4000 tokens, we spent around \$2.4 for each problem on \texttt{OpenAI} APIs if we use \texttt{gpt-4} and \texttt{text-davinci-003}, and the cost would decrease to \$0.8 if we use \texttt{gpt-3.5-turbo}. 
Notice that these estimates are computed based on the prices as of 05/14/2023, and we expect the price to further decrease in the future.
We ran the Flan-T5 based validator for \~ 2 hours on 1 80G A100 GPUs. 

The total amount of computational resources spent in this research paper is around \$2,500 in terms of \texttt{OpenAI} API and 3,000 hours of compute on A100 GPU with 80G memory.  

\section{Generation Process of \syndataset{}} \label{app:synd5}
The high-level description is in Section \ref{sec:synd5}. Here we discuss the procedure that generated \syndataset{}. 

We consider three dimensions of differences: topic, genre, and style.
For each, we generated 14/9/7 values, e.g., ``\textit{celebrity love stories}'' and ``\textit{sports team recruiting athletes} for the topic attribute, ``\textit{rap lyrics}'' and ``\textit{screen play}'' for the style attribute, and ``\textit{French}'' and ``\textit{Spanish}'' for the language attribute. 
We then used GPT-4 and the Claude API to synthesize 54K text samples, where for each text sample we sampled a topic, genre, and style randomly, e.g. ``\textit{Write a rap about a sports team recruiting athletes in French}''.
To synthesize a random \syndataset{} problem, we randomly sampled a distractor dimension (e.g. language) and a target dimension (e.g. topic), and for each dimension we sampled two random values (e.g. English and French for language, sports and art for topic). 

For each problem, we sampled 10 texts for corpus $A$ such that all of them satisfy one sampled value for the distractor dimension (e.g. corpus $A$ is entirely in English), and 10 texts for corpus $B$ for to satsify the other distractor dimension (e.g. corpus $B$ is entirely in French). 
Then we set $V$ fraction of corpus $A$ to satisfy the reference target attribute, e.g. ``is sports-related'', and $f$ fraction of corpus $B$ to satisfy the other value for the target dimension (e.g. ``is art-related'').
We chose $V$ uniformly at random from [0.6, 0.8, 1].
Finally, we provide $k$ example hypotheses from the target dimension other than the target dimension values for Corpus A and Corpus B, and we chose $k$ from [0, 2] uniformly at random.
We then sampled 300 \taskname{} problems in total from this distribution.

\section{Robustness Checks for Results on \syndataset{}} \label{app:robustness-check-similarity}

Table \ref{tab:text-davinci-003} shows the accuracy of different systems using \texttt{text-davinci-003} as the judge for semantic equivalence.
Table \ref{tab:synd5-more} shows the accuracy of different systems if we consider outputs semantically similar to the reference to be correct. 
Across all setups, we found that the conclusion reached in Section \ref{sec:quantitative} still holds under these robustness checks.

\begin{table*}[h]
    \centering
    \begin{tabular}{c|cc||c|ccc}
       \texttt{text-davinci-003}  & w/\phantom{o} goal & wo/ goal & \texttt{gpt-4} & w/\phantom{o} goal& wo/ goal\\
       \hline
       w/\phantom{o} validator  & \phantom{0}6\% & \phantom{0}1\% & w/\phantom{o} validator& 23\% & \phantom{0}9\% \\
       wo/ validator & \phantom{0}3\% & \phantom{0}0\% & wo/ validator& \phantom{0}6\% &\phantom{0}2\%\\
       \hline
    \end{tabular}
    \caption{Same Table as \ref{tab:synd5-table}, except that we use \texttt{text-davinc-003} instead \texttt{Claude-v1.3} to judge similarity. Using the validator, the goals, and \texttt{gpt-4} leads to better results.
    }
    \label{tab:text-davinci-003}
\end{table*}

\begin{table*}[h]
    \centering
    \begin{tabular}{c|cc||c|ccc}
       \texttt{text-davinci-003}  & w/\phantom{o} goal & wo/ goal & \texttt{gpt-4} & w/\phantom{o} goal& wo/ goal\\
       \hline
       w/\phantom{o} validator  & 46\% & 23\% & w/\phantom{o} validator& 53\% & 43\% \\
       wo/ validator & 24\% & 16\% & wo/ validator& 24\% & 24\%\\
       \hline
    \end{tabular}
    \caption{Same Table as \ref{tab:synd5-table}, except that we calculate how often the output is similar, rather than equivalent, to the reference. Using the validator, the goals, and \texttt{gpt-4} leads to better results.
    }
    \label{tab:synd5-more}
\end{table*}

To improve the accessibility of our research, we ran the same experiments with \texttt{gpt-3.5-turbo} and \texttt{flan-t5-xxl}, and report the results in Appendix Table \ref{tab:accessible}.
To show that our conclusions are general and not only apply to synthetically generated texts, we additionally constructed an extension of \syndataset{} with human-written texts by adapting the NYT dataset from \citet{wang2023goal}, where each text sample is a New York Times article with a topic and a location label: the topic dimension has 9 different values (e.g., politics, arts) and the location dimension has 10 different values (French, Italy);
we then followed the same procedure described in Section \ref{sec:synd5} to create this extension of \syndataset{}, and report our systems' performance in Appendix Table \ref{tab:nyt}.
Under all experimental setups, using the validator and the goal improves the performance.

\begin{table*}[h!]
    \centering
    \begin{tabular}{l|llll}
         & w/ g, w/ v & wo/ g, w/ v & wo/ g, w/v & wo/ g, wo /v \\
     \hline
     \texttt{flan-t5-xxl}  & 0.05 & 0.03 & 0.02 & 0.01 \\
     \texttt{gpt-3.5-turbo} & 0.27 & 0.10 & 0.08 & 0.03 \\
     \texttt{gpt-4} & 0.27 & 0.15 & 0.08 & 0.05 \\
     \hline
    \end{tabular}
    \caption{Similar to Table \ref{tab:quantitative}, we used \texttt{gpt-3.5-turbo} and \texttt{flan-t5-xxl} as the proposer to tackle the SynD5 dataset, and report the performance with/without using the goal (g), and with/without using the vadliator (v). We find that using the goal and the validator significantly improves the performance, and open-sourced models lag significantly behind. Additionally, \texttt{gpt-4} does not significantly outperform \texttt{gpt-3.5-turbo}.}
    \label{tab:accessible}
\end{table*}

\begin{table*}[h!]
    \centering
    \begin{tabular}{l|llll}
         & w/ g, w/ v & wo/ g, w/ v & wo/ g, w/v & wo/ g, wo /v \\
     \hline
     \texttt{gpt-3.5-turbo} & 0.61 & 0.24 & 0.22 & 0.10 \\
     \texttt{gpt-4} & 0.55 & 0.28 & 0.22 & 0.16 \\
     \hline
     \end{tabular}
    \caption{We created an extension of SynD5 by adapting a dataset of New York Times articles with two dimensions: topic and locations, each with 9 and 10 values. We then used \texttt{gpt-3.5-turbo} and \texttt{gpt-4} as the proposer, and found the same conclusion: using the goal and a validator improves the performance. Additionally, \texttt{gpt-4} does not significantly outperform \texttt{gpt-3.5-turbo}.}
    \label{tab:nyt}
\end{table*}

\section{Computing Turker Judgement} \label{app:turker}

\noindent\textbf{Scoring.} To estimate $T(h, x)$ with Turker's rating, where $h$ is a truth predicate of a text sample $x$, the Turker needs to read $h$ and $x$ and then choose among six options: ``Certainly Yes'', ``Likely Yes'',  ``Neutral'', ``Likely No'', ``Certainly No'', and ``Confusing/Cannot be decided.'' 
For each $(h, x)$ pair, we collect responses from three Turkers.
To compute the average across them, we collect a list of scores using the following rule:
each ``Certainly Yes'' would receive a score of 1.00, ``Likely Yes'' 0.75, ``Neutral'' 0.50, ``Likely No'' 0.25, ``Certainly No'' 0.00, and ``Confusing/Cannot be decided.'' receive two scores of 0.50.
We then take the average over all the scores we collected from the Turkers for one $h$ and $x$.
``Confusing/Cannot be decided.'' receives two scores of 0.50 because we want such a response to drag the average rating towards neutral and it has a larger effect than choosing ``Neutral''.

\noindent\textbf{Payment.} 
We adjust the payment for each HIT task based on the number of words they need to read. 
We pay them approximately 0.001 cent per word, and using the conservative estimate that adults read about 200 words per minute, we pay them around \$12 per hour.
We spent in total around \$5K on this HIT task.

\noindent\textbf{Qualification.}
We only recruited Turkers who are located in the U.S. Additionally, we designed qualification test with 8 questions; the questions are designed to be easy to answer as long as they have read our instructions below, and we only accepted turkers who made mistakes on at most one questions. 

\noindent\textbf{Annotation Instruction.} 
We show our annotation instruction below.
We only show examples of choosing ``Certainly Yes'', ``Certainly No'', and ``Confusing'' to encourage the Turkers not to choose neutral ratings. 
Additionally, we explicitly tried to address Halo effect -- where the text does not satisfy a predicate $h$ but satisfies a predicate $h'$ that is highly correlated with $h$.
For example, for the text sample $x=$ ``\textit{Really love the flight!!}'' does not satisfy the predicate $h=$ ``\textit{mentions that the breakfeast is good on the plane}'', even though it satisfies a highly correlated predicate $h'=$ ``\textit{likes the flight.}''

\subsection{Instructions}
Below are the same instructions we have shown you during the qualification. Thanks for visiting this page and refresh your memory about the instruction!

\textbf{Instruction}: In this task, you will check whether a TEXT satisfies a PROPERTY

\textbf{Example 1}\\
\textbf{Property}: mentions a natural scene.\\
\textbf{Text}: I love the way the sun sets in the evening.
\begin{itemize}
    \item A) Certainly Yes.
    \item B) Likely Yes.
    \item C) Neutral.
    \item D) Likely No.
    \item E) Certainly No.
    \item F) Confusing/Cannot be decided.
\end{itemize}

\noindent\textbf{Answer.} A. sun set is nature-related; if you feel a bit ambivalent, B is also acceptable.

\textbf{Example 2}\\
\textbf{Property}: writes in a 1st person perspective.\\
\textbf{Text}: Makima is cute.
\begin{itemize}
    \item A) Certainly Yes.
    \item B) Likely Yes.
    \item C) Neutral.
    \item D) Likely No.
    \item E) Certainly No.
    \item F) Confusing/Cannot be decided.
\end{itemize}

\noindent\textbf{Answer.} E. This text is undoubtedly written in the 3rd person perspetive, so E.

\textbf{Example 3}\\
\textbf{Property}: is better than group B.\\
\textbf{Text}: I also need to buy a chair.
\begin{itemize}
    \item A) Certainly Yes.
    \item B) Likely Yes.
    \item C) Neutral.
    \item D) Likely No.
    \item E) Certainly No.
    \item F) Confusing/Cannot be decided.
\end{itemize}

\noindent\textbf{Answer.} F. It is unclear what the hypothesis mean (e.g., what does group B mean?) and doesn't seem related to the text. So F.

\textbf{Example 4}\\
\textbf{Property}: mentions that the breakfast is good on the airline.\\
\textbf{Text}: The airline staff was really nice! Enjoyable flight.
\begin{itemize}
    \item A) Certainly Yes.
    \item B) Likely Yes.
    \item C) Neutral.
    \item D) Likely No.
    \item E) Certainly No.
    \item F) Confusing/Cannot be decided.
\end{itemize}

\noindent\textbf{Answer.} E. Although the text appreciates the flight experience, it DOES NOT mention about the breakfast. So the answer is E.

\textbf{Example 5}\\
\textbf{Property}: appreciates the writing style of the author.\\
\textbf{Text}: The paper absolutely sucks because its underlying logic is wrong. However, the presentation of the paper is clear and the use of language is really impressive.
\begin{itemize}
    \item A) Certainly Yes.
    \item B) Likely Yes.
    \item C) Neutral.
    \item D) Likely No.
    \item E) Certainly No.
    \item F) Confusing/Cannot be decided.
\end{itemize}

\noindent\textbf{Answer.} A. Although the text dislikes the paper, it DOES like the writing style. So the answer is A.

\section{Prompt to Judge Predicate Similarity} \label{app:similarity}
We prompt Claude v1.3 \citep{bai2022constitutional} to judge whether the predicated predicate is similar to the reference. We consider a response that leads to a``yes'' to be correct when we require the discovery to be semantically equivalent to the reference, and consider a response that leads to a ``yes'' or ``related'' to be correct when we require the discovery to be semantically similar to the reference.
\begin{quote}
``
\textit{Is text\_a and text\_b similar in meaning? respond with yes, related, or no.\\
\\
Here are a few examples.\\
Example 1:\\
text\_a: has a topic of protecting the environment\\
text\_b: has a topic of environmental protection\\ and sustainability\\
output: yes \\
\\
Example 2:\\
text\_a: has a language of German\\
text\_b: has a language of Deutsch\\
output: yes\\
\\
Example 3:\\
text\_a: has a topic of the relation between political figures \\
text\_b: has a topic of international diplomacy\\
output: related \\
\\
Example 4:\\
text\_a: has a topic of the sports\\
text\_b: has a topic of sports team recruiting new members\\
output: related\\
\\
Example 5:\\
text\_a: has a named language of Korean\\
text\_b: uses archaic and poetic diction\\
output: no\\
\\
Example 6:\\
text\_a: has a named language of Korean\\
text\_b: has a named language of Japanese\\
output: no\\
\\
Target:\\
text\_a: \{\textbf{predicate}\}\\
text\_b: \{\textbf{reference}\}\\
output:}''
\end{quote}

\section{Relevance Rating with External Non-Authors}

To make sure that the conclusion that ``using goal in the context can improve hypotheses relevance'' can be robustly reproduced by external non-authors, we also evaluated the relevance of the hypotheses with Amazon Mechanical Turks, \texttt{GPT-3.5-turbo}, \texttt{Claude-v1.3}, and \texttt{GPT-4}.
We report the results in Table \ref{tab:external-relevance} and found that the conclusion still robustly holds. 

\begin{table*}[h!]
    \centering
    \begin{tabular}{l|llll}
        Relevance Rater & w goal & w /o goal & $p$-value & spearmanr   \\
        \hline
        Authors & 1.68 & 1.20 & 1 $\times 10^{-10}$ & 1.00 \\ 
        Turkers & 1.56 & 1.44 & 4 $\times 10^{-2}$ & 0.10 \\
        \texttt{gpt-3.5-turbo} & 1.05& 0.94& 5 $\times 10^{-2}$&  0.19\\
        \texttt{claude-v1.3} & 1.18 & 0.92 & 2 $\times 10^{-3}$ &  0.30\\
        \texttt{gpt-4} & 1.49 & 1.12 & 1 $\times 10^{-6}$ & 0.45\\
    \hline
    \end{tabular}
    \caption{We rated the relevance in the same way as Table \ref{tab:rel}. However, in this table we obtained the ratings not from the authors, but from four different evaluator types: Turkers, \texttt{gpt-3.5-turbo}, \texttt{claude-v1.3} and \texttt{gpt-4}. For each evaluator type, we calculate (1) the average rating of the candidate discovery when goal is (not) present in the proposers' prompt, (2) the $p$-value that the average rating when goal is present is higher under a t-test, and (3) the spearman rank correlation between its rating and the authors' rating. We find that the $p$-value is smaller than 0.05 in all cases, indicating that our conclusion is robust; additionally, more capable models has a higher correlation with the authors.}
    \label{tab:external-relevance}
\end{table*}

\section{Meaningfulness: Relevance, Novelty, and Significance} \label{app:more-metrics}

Not every valid discovery is meaningful. 
For example, if the goal is to understand the topical differences between news from 2008 (Corpus A) and news from 2007 (Corpus B), the discovery that Corpus A ``\textit{contains news from 2008}'' is completely valid by definition but meaningless, since it provides only trivial information and is irrelevant to the goal of understanding topical differences.

\citet{mcgarry2005survey} surveyed a list of desirable properties for discovery, and we condensed them into three submetrics to rate how meaningful a discovery is based on the exploration goal: 1) relevance, 2) novelty, and 3) significance.
We evaluate these independently of validity and assume that the discovery is already valid.
For example, the discovery that ``something can travel faster than light'' is meaningful if true, even though it is highly implausible. 

We rate each submetric with \circled{0}, \circled{1}, or \circled{2}, where higher is better. 
We show the evaluation instructions below and present our rating on \texttt{text-davinci-003} proposed hypotheses.

\subsection{Evaluation Instructions}

\noindent\textbf{Relevance.}
How relevant the discovery is to the goal.
For example, suppose we were a student comparing essays rated as convincing vs. not convincing to figure out what writing style is convincing. Then: 
\begin{itemize}
    \item The discovery ``\textit{write in first person}'' is directly related to the writing style, so we rate it \circled{2}.
    \item The discovery ``\textit{use the word “I”}'', is not exactly a writing style, but can still inform the relevant underlying principle of ``\textit{write in first person}'', so we rate it \circled{1}.
    \item The discovery ``\textit{argue for abortion}'' does not tell us about the underlying writing style, so we rate it \circled{0}. 
\end{itemize}

\noindent\textbf{Novelty.}
The difficulty of generating the discovery, e.g. can we think of the discovery in 5 minutes with the goal but without looking at the corpora?
For example, suppose we were an airline manager trying to find improvements to the flight experience, and we were comparing negative reviews vs.~positive reviews. Then:
\begin{itemize}
    \item The discovery ``\textit{contain more negative language}'' is almost certain for negative reviews, so we rate it \circled{0}.
    \item The discovery ``\textit{complain about the crew members}'' is not entirely novel, but is not tautologically true and hence requires confirmation, so we rate it \circled{1}.
    \item The discovery ``\textit{mention a language barrier with the crew members}'' is specific and hard to think of without looking at the data, so we rate it \circled{2}. 
\end{itemize}

Note that our evaluation is ``blinded to the samples'': we still consider a discovery novel as long as it is hard to think of before looking at the corpora, even if it might be easy to think of after looking at the corpora. 
For example, the physical law that $F=ma$ is easy to observe if we have collected and plotted the data on acceleration, mass, and force; however, it might be difficult to think of before we see any such data, so we consider it novel. 

\noindent\textbf{Significance.}
Given the exploration goal, how beneficial is it to learn the discovery for the first time?
For example, suppose we were an Amazon retailer trying to figure out what customers like and dislike about my product based on negative reviews and positive reviews. Then:
\begin{itemize}
    \item The discovery ``\textit{accuses the team pushing out a bad product}'' is not significant since it cannot direct the retailer to improve the product, so we rate it \circled{0}. 
    \item The discovery ``\textit{asks for a more durable product}'' gives some hints about how to improve the product, but isn’t sufficiently helpful on its own, so we rate it \circled{1}. 
    \item The discovery ``\textit{says the wrench is missing}'' can lead to concrete actions for improvement, so we rate it \circled{2}. 
\end{itemize}

\subsection{Goal Leads to More Meaningful Hypotheses}

\begin{table*}[h!]
    \centering
\begin{tabular}{lcc|cc|cc}
\hline
{} &  with-goal &  no-goal &  kappa &  spearmanr  & $p$ of avg & worst $p$ of ind\\
\hline
Relevance &          1.68 &             1.20 &   0.56 &       0.71 & 1 $\times$ 10$^{-10}$ & 1 $\times$ 10$^{-8}$\\
Novelty &          1.24 &             0.97 &   0.37 &       0.50 & 5 $\times$ 10$^{-6\phantom{0}}$ & 4 $\times$ 10$^{-2}$ \\
Significance &          1.56 &             1.05 &   0.46 &       0.64 & 2 $\times$ 10$^{-10}$ & 2 $\times$ 10$^{-7}$\\
\hline
\end{tabular}
    \caption{\textbf{Left.} For each metric, we report the average rating on hypotheses generated with or without using the exploration goal, and find that the former performs better. \textbf{Middle.} The inter-annotator agreement rate averaged across pairs of author evaluators, measured by Kappa and Spearman rank coefficient; we find substantial correlations between evaluators across all these subjective metrics, with relevance $>$ significance $>$ novelty. 
    \textbf{Right.} 
    We compute the $p$-values for the null hypothesis that ``with-goal and no-goal result in the same performance''. The $p$ of avg column reports the $p$-values after we average the ratings from all evaluators, while the ``worst $p$ of ind'' column takes the max of all $p$-values based on ratings of individual evaluators. 
    Overall, the conclusions are statistically significant and they can be robustly reproduced across individual evaluators.
    }
    \label{tab:quantitative}
\end{table*}

Compared to \citet{zhong2022describing}, we added the exploration goal to our prompt when generating hypotheses.
Does this improve the quality of the proposed hypotheses?
To investigate this, we sampled 100 problems from \datasetname\phantom{0}with distinct exploration goals and randomly sampled 2 hypotheses from GPT-3 with and without using exploration goal (see Figure \ref{fig:prompts}), resulting in 400 hypotheses to evaluate.
Three authors then rated their meaningfulness based on the three metrics defined in Section \ref{sec:metrics}, while being blinded about which hypotheses were generated with the exploration goal.

The results are shown in Table \ref{tab:quantitative}. 
We found that, when prompted with the exploration goal, GPT-3 on average proposes more relevant, novel, and significant hypotheses;
additionally, it proposes hypotheses with ratings higher than \circled{0} 31\%/21\%/28\% more often in terms of relevance/novelty/significance.
Since this is a subjective evaluation, the Kappa inter-annotator agreement is only moderate, ranging from 0.37 to 0.56.
However, we can still robustly conclude that the model can propose more meaningful hypotheses when conditioned on the goal: we calculate the $p$-values for the null hypothesis that with-goal and no-goal have equal performance, and we find $p$-values to be highly significant and robust across evaluators, for all three submetrics.

\section{Full Pipeline of the Proposer} \label{app:proposer}

We present the full details of how we generated the hypotheses with the language model.
The process roughly contains four stages: 1) obtaining representative samples for each corpus, 2) sampling hypotheses from GPT-3, 3) rewriting hypotheses, and 4) optionally plugging in example hypotheses.

\paragraph{Obtaining representative samples.} 
This step is the same as \citet{zhong2022describing}, and we borrow the related text from that paper for the reader's convenience.
Since $\mathcal{D}^{\text{res}}_{A}$ and $\mathcal{D}^{\text{res}}_{B}$ might overlap significantly, random samples from $\mathcal{D}^{\text{res}}_{A}$ and $\mathcal{D}^{\text{res}}_{B}$ might not be representative and informative enough for GPT-3 to notice the differences between the two distributions.
Therefore, we choose samples that are representative of their differences.
To find those samples, we fine-tune RoBERTa-Large 
\cite{liu2019roberta} to predict whether each sample comes from Corpus A or Corpus B and keep the top-p percentile samples with the highest confidence.
Next, we take samples from the top-$p$ percentile to prompt GPT-3.

\paragraph{Selecting samples to prompt GPT-3.}
We randomly select $S=$25 samples from the top-5 percentile from Corpus A and Corpus B to prompt GPT-3 to propose the hypotheses, using the template shown in Figure \ref{fig:prompts} left.
We require the length of the prompt to be at most 3,200 GPT-3 tokens (the max window size for GPT-3 text-davinci-003 is 4096) and gradually decrease the number of samples $S$ in the prompt until the prompt length is less than 3,200;
additionally, we truncate each text samples to at most 256 GPT-3 tokens.
Finally, to prevent GPT-3 from proposing hypotheses that reflect simple lexical correlations that can be detected with unigram models, e.g., ``\textit{uses the word ``hey'' more often.}'', we incrementally construct the subset of samples for Corpus A and Corpus B such that at any time of the construction, no single word can appear $0.25S$ times more often in one corpus than the other. 
We repeat the same process for the top-20 and top-100 percentile until we obtain 60 hypotheses.

\paragraph{Rewriting hypotheses with GPT-3.}
As mentioned in Section \ref{sec:analysis},  the hypotheses generated by GPT-3 are frequently statements about the corpus, while the validator requires the hypothesis to be a predicate on individual text samples.
For example, when comparing definitions that people like from \url{UrbanDictionary.com} to other definitions, the hypothesis that the former ``\textit{is more likely to include slang or colloquial terms.}'' is a statement about a collection of text samples, rather than a predicate on an individual sample.
$T(h, x)$ is undefined in this case, since it does not make sense to check whether a single text sample is more likely to include slang.
Ideally, we want to detect these comparison statements and automatically remove the comparatives, e.g., rewrite it to ``\textit{includes slang or colloquial terms.}''.

\begin{figure}
    \centering
    \includegraphics[width=0.5\linewidth]{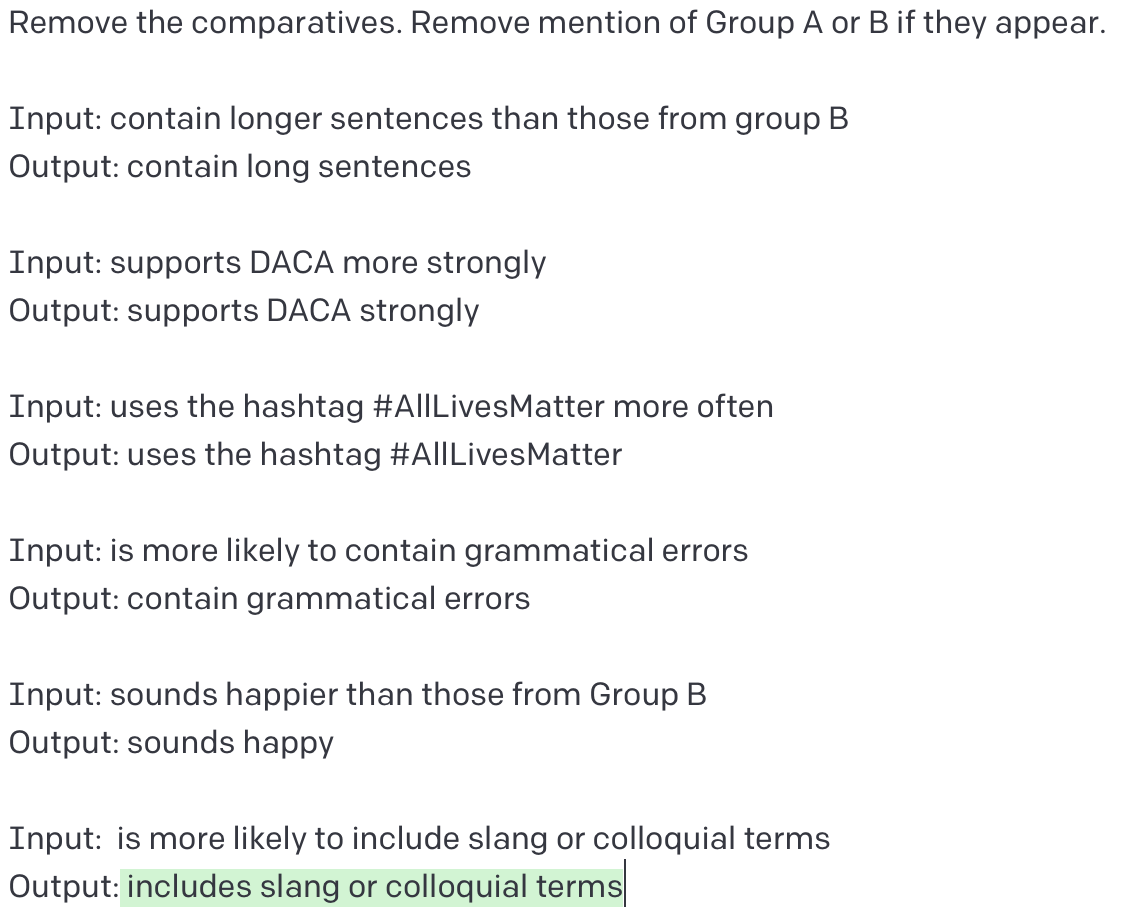}
    \caption{The prompt to remove comparatives from a hypotheses.}
    \label{fig:rm-cmp}
\end{figure}

To detect and remove the comparatives from the hypotheses, we tag the part of speech for each word in the hypotheses using the NLTK package \citep{bird2009natural} and check whether any tag is \texttt{JJR} or \texttt{RBR}. 
If a hypothesis indeed contain theses tags, we prompt GPT-3 to rewrite the hypothesis. 
We show an example prompt in Figure \ref{fig:rm-cmp}.

\paragraph{Plugging in example hypotheses (optionally).}
We can also add a few problem-specific example hypotheses to the prompt to elicit more relevant hypotheses, and we do so by adding them to the ``formatting instruction'' part in the prompt used to propose hypotheses Figure \ref{fig:prompts}.
In \datasetname{}, we provided example hypotheses for each problem to steer our system to generate more meaningful discoveries; we produced the example hypotheses by prompting GPT-3 to generate a few hypotheses and selecting the meaningful ones from them.

For the reported discoveries in Section \ref{sec:application}, we confirmed that they are unambiguously different from our provided hypotheses; otherwise, the system might have produced the discoveries by copying the provided hypotheses.
We did not use the example hypotheses in Section \ref{sec:quantitative} to test GPT-3's zero-shot understanding of the goal.

\section{Collecting Data to Fine-tune the Validator} \label{app:verifier}

Here we provide a high-level description of how the data was collected. 
For each problem in \datasetname{}, we used our proposer to produce a list of hypotheses.
We automatically judged each hypothesis on a subset of samples from the research split using GPT-3 text-davinci-002 \citep{ouyang2022training}, Flan-T5 \citep{chung2022scaling}, and a model trained with RLHF from \citet{bai2022training}. 
We created the input distribution for training by combining and equally weighting the following $3 \times 2 = 4$ distributions: the subset of $(h, x)$ pairs that GPT-3/Flan-T5/``RLHF'' considers Yes or No to be the most likely answer.
We then collected averaged turker ratings for in total 3138 $(h, x)$ pairs and used them to fine-tune Flan-T5 to create the validator \citep{chung2022scaling}.

To test cross problem generalization capability of our \taskname{} system, whenever we applied our \taskname{} system to a problem in \datasetname{} in Section \ref{sec:application}, we used a validator that is NOT fine-tuned on the $(h, x)$ pairs from this problem.
We achieved this by keeping track of which problem each $(h, x)$ pair comes from and split all the $(h,x)$ pairs into three folds based on the problems; whenever we applied our \taskname{} system to a problem, we used the validator trained on the two folds that do not contain this problem.

\section{What Discoveries Did we Choose to Present} \label{app:choice}
Our system in total produces 3296 discoveries on \datasetname{}. 
However, we do not have enough budget to validate every finding, since estimating $V$ is expensive (Section \ref{sec:soundness}).
Therefore, from the remaining 3296 discoveries, we manually selected 21 discoveries that 1) the authors think are relevant enough, 2) are representative of potential use cases, 3) do not require expert knowledge for Turkers to judge, and 4) are likely to achieve a small $p$-value with fewer than 200 samples from $\mathcal{D}^{\text{val}}_{A}$ and $\mathcal{D}^{\text{val}}_{B}$. 
We then estimated their validity based on the procedure described in Section \ref{sec:soundness} by using fewer than 200 samples from the validation split and calculated the $p$-values.\footnote{We determined the number of samples s.t. $V'$ can achieve a $p$-value of $0.005$. Estimating $V$ for these discoveries costs $\sim$\$1500.} 
Since we are testing multiple discoveries and each of them can be statistically significant merely due to chance, we keep 13 discoveries with $V$ that are significantly non-zero with $p$-value below 7\%, a threshold determined by the Benjamini Hochberg's procedure with a false discovery rate of 10\%.
In other words, fewer than 10\% of the discoveries presented are false discoveries in expectation.

\section{More Example Discoveries on \datasetname{}} \label{app:more}
\begin{table}[]
    \centering

\begin{tabular}{lllll}
\toprule
                                         discovery &     V &         p &    V' &         p' \\
\midrule
 argues for a path forward to promote the fair ... &  0.16 &  1.26e-04 &  0.35 &   2.01e-73 \\
         refers to illegal immigrants as criminals &  0.09 &  6.17e-03 &  0.19 &   3.17e-38 \\
 has an informal tone, such as slang or colloqu... &  0.08 &  2.35e-03 &  0.24 &   1.46e-35 \\
                          mentions lack of legroom &  0.16 &  1.15e-03 &  0.38 &   1.34e-45 \\
                       mentions children or family &  0.08 &  1.00e-05 &  0.11 &   8.05e-09 \\
       Uses language that is positive or uplifting &  0.12 &  2.12e-03 &  0.24 &   4.18e-59 \\
                 references violence or aggression &  0.06 &  9.87e-03 &  0.17 &   4.25e-26 \\
 involves physical activity, such as walking, p... &  0.13 &  4.92e-03 &  0.37 &  7.07e-101 \\
 contains keywords related to business, finance... &  0.08 &  2.89e-02 &  0.35 &   1.45e-95 \\
 mention disasters and crimes, such as plane ac... &  0.03 &  7.03e-02 &  0.09 &   4.61e-06 \\
              discusses coronavirus-related topics &  0.21 &  1.01e-04 &  0.27 &   9.19e-78 \\
 references pop culture, such as movies, books,... &  0.21 &  2.67e-04 &  0.58 &   2.09e-30 \\
 uses vivid imagery and metaphors to convey a f... &  0.09 &  2.47e-02 &  0.45 &   5.04e-64 \\
\bottomrule
\end{tabular}
    \caption{The full table of discoveries, along with their $V$, $V'$, $p$, and $p'$ scores.}
    \label{tab:full-discovery-values}
\end{table}

\noindent\textbf{Analyzing errors in NLP systems.}
We considered the task of perspectrum classification \citep{CKYCR19}, which has the following instruction: ``given a perspective and a claim, classify whether the given perspective supports or undermines the claim. If the perspective could possibly convince someone with different view, it is supporting, otherwise it is undermining.''
We considered two few-shot learning systems: GPT-3 Instruct Curie \citep{ouyang2022training} and Tk-Instruct-11B \citep{wang2022super}.
We focused on the perspectives where the ground truth label is undermining, and compare the following two corpora: Corpus A -- the set of perspectives where Curie correctly classifies the input as undermining but Tk-11B is wrong, and Corpus B -- the set where TK-11B is correct while Curie is wrong. 
We found that Corpus B more often ``\textit{Uses language that is positive or uplifting}'' ($V \approx 0.12$, AUCROC $\approx$0.67). 
One possible explanation is that Curie made many mistakes by misinterpreting undermining as a label for negative sentiment rather than a logical relation between the claim and the perspective. 

\noindent\textbf{Comparing lyrics from different eras.}
Compared to lyrics from the 70s, those from the 80s more often ``\textit{references violence or aggression}'' ($V \approx 0.06$, AUCROC $\approx$ 0.58).

\noindent\textbf{Describing distribution shift.}
We compared the premises from the SNLI dataset and MNLI dataset, and the former ``\textit{involves physical activity, such as walking, playing, climbing, or biking}'' ($V \approx 0.13$, AUC-ROC $\approx$0.64).
One possible explanation is that SNLI is based on image captions.

\noindent\textbf{Comparing discussion topics between bots and human users.}
We compared the topical differences between tweets identified as written by bots vs. human users on Twitter, and our system finds that the bots more often ``\textit{contains keywords related to business, finance or trading}'' ($V \approx 0.08$, AUC-ROC $\approx$ 0.61). 
One possible explanation is that bots are frequently used to generate finance-related scams.

\noindent\textbf{Identifying temporal differences in news headlines.}
We compared headlines published by ABC news across different years. 
Compared to 2014, headlines from 2010 ``\textit{mention disasters and crimes, such as plane accidents and assaults}'' more often ($V \approx 0.03$, AUCROC $\approx$ 0.53). Compared to year 2019, year 2020 more often ``\textit{discusses coronavirus-related topics}'' ($V \approx 0.21$, AUCROC $\approx$ 0.65).

\noindent\textbf{Describing text clusters.}
We present two example descriptions for text clusters. 
One from Wikipedia: ``\textit{references pop culture, such as movies, books, and television shows}.'' ($V \approx 0.21$, AUC-ROC $\approx$ 0.73); 
one from PoetryFoundation.com: ``\textit{uses vivid imagery and metaphors to convey a feeling}'' ($V \approx 0.09$, AUC-ROC $\approx$0.65).

\section{Limitations and Future Work} \label{app:limit}
We still face many challenges in building a broadly useful system.
We describe technical challenges that machine learning researchers can tackle in Appendix \ref{app:engineering-challenges} and organizational challenges that require domain experts in Appendix \ref{app:organizational-challenges}.

\subsection{Engineering Challenges} \label{app:engineering-challenges}

\noindent\textbf{Hypotheses about the corpora might not be appropriate predicates on individual samples.}
When comparing highly rated definitions from \url{UrbanDictionary.com} to others, our system generates the hypothesis that the former ``\textit{is more likely to include slang or colloquial terms.}'' This is a statement about a collection of text samples, but the validator requires the hypothesis $h$ to be a predicate on individual text samples $x$.
To address this, we used GPT-3 to automatically remove comparatives from the hypotheses, e.g. rewriting the hypothesis above to ``\textit{include slang or colloquial terms}.''

However, some versions of this problem were harder to remove.
For example, when comparing reviews from American Airlines (AA) flights and Delta Airlines to understand which aspects of each airline are doing better/worse, the proposer generated the hypothesis ``\textit{mentions \textbf{American Airlines'} staff being unfriendly and unhelpful}''.
Interpreted literally, this hypothesis can only be true on the corpus of AA reviews, since it presupposes the review to be about AA. The correct predicate for use on individual samples should instead be ``\textit{mentions staff being unfriendly and unhelpful}'' (without the words ``\textit{American Airlines}''').
Therefore, future systems should explicitly convert corpus-level statements to their corresponding correct predicates, and the metrics should evaluate whether the validity of the predicates implies the corpus-level statements.

\noindent\textbf{Beyond truth predicates.} 
Our work requires the discovery to be a truth predicate that maps a text sample to a truth value. 
However, scientific discoveries can be arbitrary natural language expressions; extending to more flexible expressions requires a significant redesign of our system and evaluation framework.
Some more feasible near-term extensions include 1) allowing natural language expressions that map from text samples to real values, e.g., ``how polite the sentence is compared to other samples from the corpora'' or 2) using additional logical forms to combine individual truth predicates; e.g., learn a shallow and interpretable decision tree where each split point is a natural language predicate.

\noindent\textbf{Beyond corpus-level differences.}
Our work focuses on describing corpus-level differences and validates a discovery by comparing how often it is true on each corpus.
Future work can consider other ways to validate a discovery: for example, suppose each text sample is associated with a continuous target variable, we can validate whether a discovery is more likely true if the target variable is large. 

\noindent\textbf{Investigating sensitivity towards prompt format.}
In this paper we hand-crafted the prompt for the proposer and manually annotated the exploration goals on our own for \datasetname{}. 
However, due to budget limitation, we have not investigated how sensitive is our D5 system towards prompt formatting and paraphrasing, or whether the performance could have been improved with better prompts.
Future works can investigate more in this research direction.

\noindent\textbf{Clarifying a discovery.}
Some discoveries seem to have clear meanings on the surface, but they become ambiguous when we judge them on individual text samples.
For example, judging whether a text sample $h=$ ``\textit{mentions people}'' seems like an unambiguous task a priori; however, it is unclear whether it is true on the sample $x=$ ``\textit{I woke up this morning.}'', since the ``\textit{people}'' in $h$ is a plural form, while $x$ only mentions one person ``\textit{I}''.
Future work can use a language model to automatically clarify the meaning of a hypothesis and make it more specific, e.g., rewrite $h$ as ``\textit{mentions one or more humans.}''

\noindent\textbf{Correlation $\neq$ causation.}
Like other tools that rely on correlations to analyze patterns in data (e.g., linear regression), our system cannot establish causal relations either.
For example, when comparing self-reported happy moments from females and males, even if the former corpus has more samples that ``\textit{mention children and family}'', it does not necessarily imply family plays a more important role in inter-personal relations for females; an alternative hypothesis is that females might mention any other people more often than males, hence leading to the observation that they mention family more often. 
Future work can use language models to propose what control hypothesis to test. 

\noindent\textbf{Decreasing the cost of validation.}
As alluded to in Section \ref{sec:metrics}, estimating $V$ is extremely expensive as it requires a lot of human labor.
Future work can consider an importance sampling procedure that uses $\hat{T}$ as a proposer to improve the sample efficiency of estimating $V$. 

\noindent\textbf{Training a better proposer.} 
We developed a self-supervised learning algorithm to propose more valid hypotheses. 
However, it does not take into account the meaningfulness metric, and it is unclear how to manage its trade-offs with validity if they exist.
We look forward to future works that can train a better proposer with as minimal supervision as possible. 

\noindent\textbf{Combining Meaningfulness and Validity Metrics.}
To simplify evaluation, we assumed meaningfulness to be independent of the magnitude validity $V$.
Such an assumption allows us to directly evaluate hypotheses that are not necessarily valid but is also limiting for evaluating the final discoveries: for example, for that 2008 ``\textit{discuss economy}'' more often than 2007, it would be way more significant if $V=0.99$ compared to $V=0.0000001$.
Future works can propose better metrics that do not assume that validity and meaningfulness are independent.

\noindent\textbf{Extending to Non-English Language}
\datasetname{} is currently annotated with English goals and most of the corpora are in English. 
Future work can consider extending this to other languages.

\subsection{Organizational Challenges}\label{app:organizational-challenges}

As discussed in \citet{polanyi2000republic}, it requires implicit community norms rather than explicit deductive logic to decide what counts as good research results;
to guide our system to produce truly important discoveries, our system needs feedback from researchers who work in the domain of interest.
However, except for machine learning, the authors do not have research expertise in most of the domains listed in Figure \ref{fig:dataset-overview}.
We look forward to future contributions from other domains and list concrete directions below.

\noindent\textbf{What problems to solve?}
We generated the problems in \datasetname{} by reading relevant papers and guessing what domain experts might care about. 
However, our guesses can be inaccurate.
Future works can directly gather problems from domain experts to reflect the actual usage of our system.

\noindent\textbf{How to interpret a discovery?}
We asked for Turker's judgment to compute $T(h, x)$.
However, many hypotheses require expert knowledge to interpret properly.
For example, only law experts can reliably judge whether a contract $x$ satisfies the predicate $h$ ``\textit{contains a license grant that is irrevocable}.''
Domain experts are needed to evaluate the validity of a discovery and supervise the validator. 

\noindent\textbf{What discoveries are meaningful?}
Our work developed the evaluation instructions to approximately evaluate what hypotheses are meaningful.
However, just as no one can become an outstanding peer reviewer simply by reading the review guideline, we do not consider it feasible to provide a gold evaluation simply by reading our instructions. 
Whether a discovery is meaningful depends heavily on implicit community norms, and we hope domain experts can provide better evaluation and training signals for our system.

\section{Self-Supervised Learning with Open-Ended Problems: A Proof of Concept} \label{app:training}
Since the problems in \datasetname{} are open-ended, our system could potentially produce discoveries with higher validity scores than our current system.
Therefore, we design a self-supervised learning algorithm to improve an LM's ability to propose more valid hypotheses, using the principle that \textbf{it is easier to validate a discovery than to generate one}.
%We show preliminary evidence that our algorithm is useful by improving an LM's ability to describe differences between groups of four samples. 

\noindent\textbf{Algorithm.}
Suppose we are given a set of problems for training and an initial language model $m_{\text{init}}$.
Our goal is to automatically generate a set of \textit{prompt}-\textit{completion} pairs to fine-tune $m_{\text{init}}$ so that it can propose hypotheses that are more valid. 
To generate a \textit{prompt}, we randomly sample a problem and create a proposer prompt following the procedure in Section \ref{sec:proposer}.
To generate the desired \textit{completion} given a prompt, we sample multiple hypotheses from $m_{\text{init}}$, approximate their $V'$ score on the samples in the proposer prompt with the same language model $m_{\text{init}}$ (Section \ref{sec:verifier}), and select the highest scoring hypothesis.
Finally, we use the prompt-completion pairs to fine-tune $m_{\text{init}}$.

However, since we cannot fine-tune instruction-tuned GPT-3, we can only experiment with Flan-T5 \citep{chung2022scaling}, an open-sourced instruction-tuned model that might only work well for easier ``mini-problems''.
As a proof of concept, we tested our algorithms for describing groups of four samples, where each group comes from a text cluster.
As an overly simplified example, we will give the LM the prompt  ``\textit{Group A: 1. dog 2. cat 3. pig 4. cow. Group B: 1. phone 2. laptop 3. desk 4. cup}'' as an input and the LM can output ``\textit{mentions an animal}'' as a hypothesis.
%of how group A differs from group B.

\noindent\textbf{Data.}
We created 33 corpora by merging all corpora in \datasetname{} with the same domain, and automatically generated 4503 text clusters using RoBERTa embeddings \citep{aharoni-goldberg-2020-unsupervised}. 
We focused on clustering because it can automatically generate a large amount of semantically coherent groups of samples.
To create a pair of four samples, we randomly sampled a corpus, sampled two clusters within that corpus, and took four random samples from each cluster.
To test cross-corpus generalization, we reserved 28 of the 33 corpora to create mini-problems for evaluation, using the rest for training.
We used Flan-T5 \citep{chung2022scaling} as $m_{\text{init}}$ and sampled hypotheses with a temperature of 0.8.
For training, we sampled 30,000 mini-problems and selected the best of eight hypotheses generated by $m_{\text{init}}$ as the target completion;
for evaluation, we sampled 200 mini-problems to calculate $V$ with Turkers and 1500 mini-problems to calculate $V'$ automatically. 

\noindent\textbf{Results.}
We evaluated randomly sampled hypotheses from the language model before and after self-supervised training.
The automated ``self-evaluation'' validity score $V'$ improves substantially from 0.22 to 0.37, and the ``true'' validity score $V$ according to Turker evaluation improves from 0.07 to 0.10, with a $p$-value of 0.02.
This result provides preliminary evidence that our algorithm (or similar variants) could be applied to a large set of problems to improve the validity of the hypotheses; we expect future validators to simulate human judgments better, hence decreasing the approximated gap of improvement between $V$ and $V'$.

\section{Comparing \taskname{} to Naïve Bayes} \label{app:bayes}
We qualitatively compare the discovery generated by our \taskname{} system to the top-5 unigram features extracted by Naive Bayes, a traditional exploratory analysis method. 
The Naive Bayes method is effective when the target difference can be saliently reflected by individual words. 
For example, ``\textit{yo}'' implies a rap genre, ``\textit{die}'' implies a language of Deutsch, and [``\textit{rank}'', ``\textit{higher}'', ``\textit{univeristy}''] hints at the topic of ``\textit{college ranking changes}''.
Additionally, compared to black-box neural networks, such a method is fully interpretable.

In comparison, \taskname{} can directly generate a semantically coherent description for the target difference, saving users' time to guess the underlying correlation by inspecting the top unigram features. 
Additionally, it can capture differences that are hard to detect at a word level; 
for example, ``\textit{the genre of biblical scripture}'' is mainly reflected in its sentence structure rather than individual words.
Finally, \taskname{} only describes goal-related differences, while Naïve Bayes picks up on any discriminative feature;
for example, when identifying the topical differences between a English and a Deutsch corpus, Naïve Bayes fails catastrophically and only picks up common determiners such as ``\textit{the}'' or ``\textit{die}'' instead of topic words, since they are the most useful feature at telling which sample comes from which corpus.
Given the respective strength of \taskname{} and traditional exploratory methods, we envision \taskname{} to serve as a complementary method to traditional methods.

\section{Annotation Interface to Collect Human-Generated Hypotheses} \label{app:interface}

(This section describes an interesting research direction we did not have time to fully pursue.)

\noindent\textbf{Task.} 
To fine-tune the language model to propose better hypotheses and perform validation more accurately, we also designed an interface to collect human annotations earlier in the project. 
In this annotation task,  the annotators see five text samples from each of the two corpora; they then write one or many natural language predicate(s) that describe how samples from the two groups are different and choose which text samples satisfy each predicate the annotator has written.
Since it is challenging for humans to identify systematic differences between even groups of five sentences, we made the task easier for them by 
\begin{itemize}
    \item we chose the representative samples from each corpus to form the two groups of samples, similar to the process in Section \ref{app:proposer}, and 
    \item we highlighted subspan of the text samples that are informative for how the two corpora differ. For example, if Corpus A is sports related while Corpus B is entertainment related, we hope to highlight sports-related words like ``basketball''. 
    To automatically identify the text spans to highlight,
    we fine-tuned RoBERTa to classify whether a sample comes from Corpus A and Corpus B, used the SHAP library to calculate how much each text span influences the classifier's decision, and highlighted the text spans based on the influence. 
\end{itemize}
\begin{figure*}[h!]
    \centering
    \includegraphics[width=\linewidth]{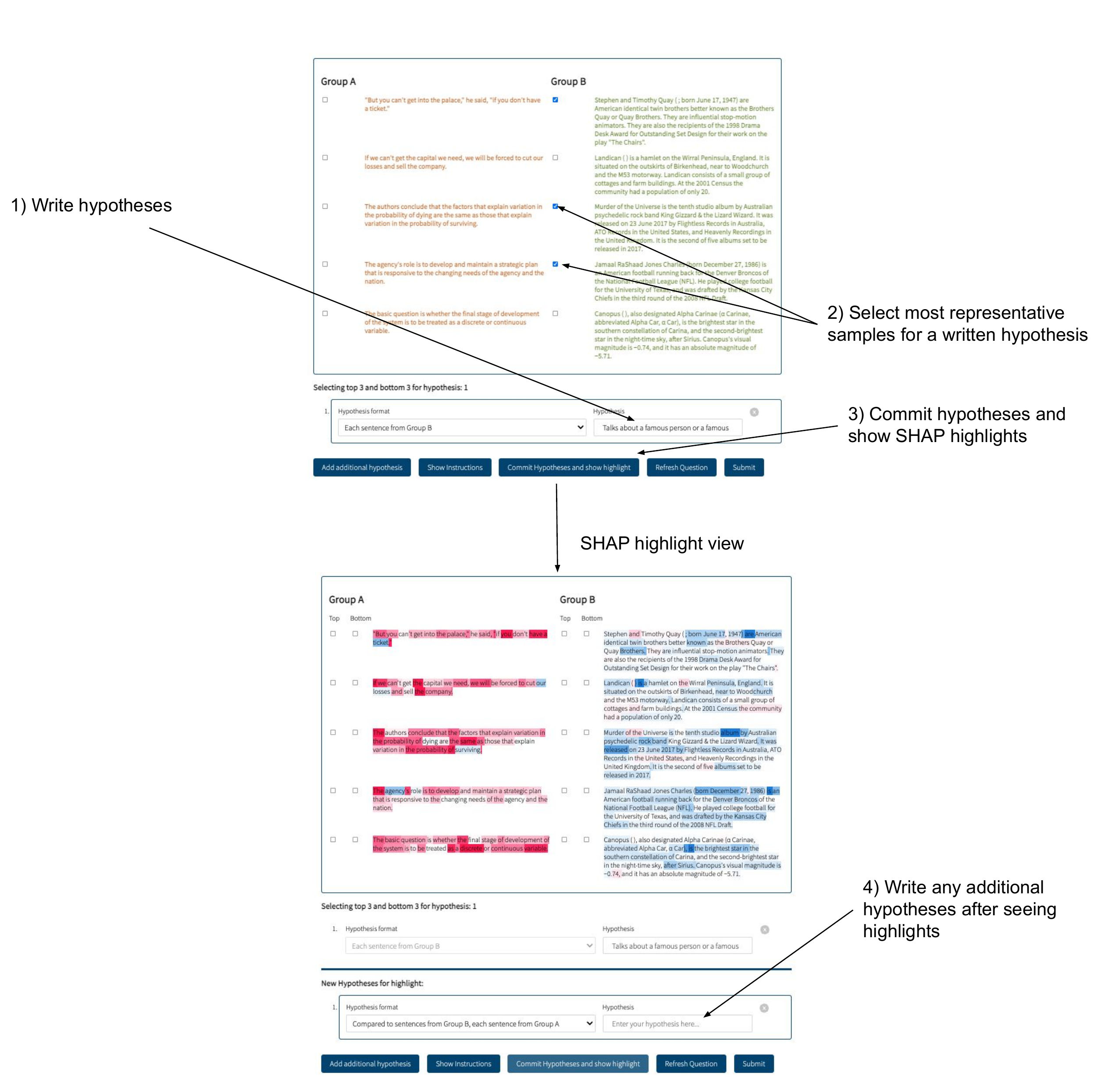}
    \caption{A detailed screenshot of our annotation interface.}
    \label{fig:survey}
\end{figure*}

A screenshot of the annotation interface can be seen in Figure \ref{fig:survey}. 

\noindent\textbf{Preliminary Results}
We performed initial experiments on text clusters formed on the wikitext-2 dataset \citep{wikitext}. 
We asked the authors to write hypotheses for 30-50 samples and then compare the results with GPT-3 generated hypotheses. 
We found that human annotators were able to write 2-4 valid hypotheses per pair of text groups, while GPT-3 text-davinci-003 was able to generate 4-6. 
Out of the valid generated hypotheses, approximately a third were variations on another valid hypothesis. 
The number of times humans were able to write a hypothesis that GPT-3 was unable to generate was around a third of the samples, while GPT-3 was able to generate a novel hypothesis humans have not thought about before in nearly every single text corpora. 
Given that GPT-3 is close to our author's ability to write hypotheses, we estimated that we would not be able to fine-tune T5 to propose better hypotheses with human annotations, and hence gave up on this research direction.

\section{Datasets} \label{sec:datasets}

Many of our datasets come from the following sources: 
the Computational Models of Social Meaning class from Columbia University\url{http://www1.cs.columbia.edu/~smara/teaching/S18/}, 
the ACL Anthology \url{https://aclanthology.org}, , and Kaggle datasets with an NLP tag. \url{https://www.kaggle.com}

\textbf{abc-headlines}. We collect headlines published by ABC news, an American news company from \citet{abc-headlines}. 
ABC headlines are directly downloaded from Harvard Dataverse. The year is extracted from the publication date field. Samples are constructed from the headline text.
The data is downloadable from \url{https://doi.org/10.7910/DVN/SYBGZL} with license CC0 1.0.

\textbf{ad-transcripts}. We collect ad scripts from a variety of industries from \citet{ad-transcripts}. Ad transcripts are directly downloaded from Kaggle. The top eight industries by frequency are selected. Newlines are replaced with spaces.
The dataset is downloadable from \url{https://www.kaggle.com/datasets/kevinhartman0/advertisement-transcripts-from-various-industries} with license CC0 Public Domain.

\textbf{admin-statements}. We collect statements of administration policy from American presidents from \citet{admin-statements}. Administration statements are extracted from a collection hosted on GitHub. Extraneous symbols are removed and samples are split by paragraph.
The dataset is downloadable from \url{https://github.com/unitedstates/statements-of-administration-policy\#statements-of-administration-policy} and origin files have a Creative Commons Attribution 3.0 License.

\textbf{ai2-natural-instruction}. We collect a learning-from-instructions dataset released by the Allen Institute for AI from \citet{ai2-natural-instruction}. Natural instruction tasks are directly downloaded without modification. The dataset is released under an Apache-2.0 license.

\textbf{airline-reviews}. We collect reviews of airlines collected from the review website Skytrax. Airline reviews for airlines, airports, and seats are downloaded from a public GitHub repository. Names of aircraft, airlines, countries, and traveler types are standardized. Ratings of 1, 4, or 5 on a scale of 5, and 1, 5, 8, or 10 on a scale of 10 are kept. This dataset can be downloaded via \url{https://github.com/quankiquanki/skytrax-reviews-dataset}.

\textbf{aita}. We collect posts on the ``Am I The Asshole'' Subreddit, an online forum people ask others whether they were in the wrong from \citet{aita}. Posts from r/AmITheAsshole are downloaded from a praw scrape of Reddit. Topic areas are chosen based on common themes in posts and coarsely defined based on manual keywords. Each post can belong to multiple topic areas. 
The dataset can be downloaded at \url{https://doi.org/10.5281/zenodo.3677563}.

\textbf{all-the-news}. We collect news articles collected from various outlets between 2015 and 2017 from \citet{all-the-news}. News articles are downloaded directly from the Components website. The titles are used as text samples.The dataset can be downloaded at \url{https://components.one/datasets/all-the-news-articles-dataset} .

\textbf{amazon-reviews}. We collect Amazon reviews collected from various product categories from \citet{amazon-reviews}. Amazon reviews are downloaded from a 2018 crawl of the website. The first 100,000 review texts are treated as the text sample.
The dataset can be downloaded at \url{https://nijianmo.github.io/amazon/index.html} .

\textbf{armenian-jobs}. We collect job postings in Armenia from \citet{armenian-jobs}. The Armenian job postings dataset is downloaded from a snapshot on GitHub. Different IT jobs are manually coded and time intervals are defined in order to balance sample availability.
The dataset can be downloaded at \url{https://www.kaggle.com/datasets/udacity/armenian-online-job-postings} .

\textbf{boolq}. We collect a reading comprehension dataset of yes/no questions from \citet{boolq}. Boolean questions are downloaded directly as is.
The dataset can be downloaded at \url{https://github.com/google-research-datasets/boolean-questions} with license CC-SA-3.0.

\textbf{clickbait-headlines}. We collect headlines across time from the Examiner, a clickbait news site from \citet{clickbait-headlines}. The Examiner headlines are directly downloaded from Kaggle. The year is extracted from the publication date field. Samples are constructed from the headline text.
The dataset can be downloaded at \url{https://www.kaggle.com/datasets/therohk/examine-the-examiner}, with license CC0: public domain.

\textbf{convincing-arguments}. We collect arguments on a variety of topics annotated for convincingness from \citet{convincing-arguments}. Annotated arguments are downloaded from the GitHub repository. Arguments are sorted by rank. The bottom 400 are treated as ``unconvincing'', the top 200 are treated as ``convincing'', and the next 200 are treated as ``somewhat convincing.''
The dataset can be downloaded at \url{https://github.com/UKPLab/acl2016-convincing-arguments}, with license CC-BY 4.0.

\textbf{craigslist-negotiations}. We collect dialogue from Craigslist negotiations, an online seller platform from \citet{craigslist-negotiations}. Craigslist negotiations are downloaded from Huggingface. Sequences which contained a ``quit'' intention or ``reject'' intention are categorized as failures; those which contained an ``accept'' intention are categorized as successes. The mid-price is defined as the mean price of the items sold. Within each category, the items are sorted by mid-price. The top half is treated as high-price and the bottom half is treated as low-price. 
This dataset can be downloaded at \url{https://huggingface.co/datasets/Hellisotherpeople/DebateSum} with MIT license.

\textbf{debate}. We collect evidence compiled for American competitive policy debate, published online by debate camps from \citet{debate}. The train split is downloaded from Huggingface. For each sample, we use the abstract as the text. Arguments are categorized by type, debate camp of origin, and topic/specific argument. For topics, we use domain knowledge to list relevant keywords for each topic and include any sample with a file name that includes any keyword. A single sample can belong to multiple topics.
This dataset can be downloaded at \url{https://huggingface.co/datasets/Hellisotherpeople/DebateSum} with MIT license.

\textbf{dice-jobs}. We collect American technology job postings on dice.com from \citet{dice-jobs}. Job postings are downloaded from Kaggle. Posts from the six most popular companies are categorized by company. We remove miscellaneous characters and blank descriptions. We additionally apply our splitting procedure to reduce description length.
This dataset can be downloaded at \url{ https://www.kaggle.com/datasets/PromptCloudHQ/us-technology-jobs-on-dicecom} under CC BY-SA 4.0 .

\textbf{diplomacy-deception}. We collect dialogue from games of Diplomacy, which involves deception from \citet{diplomacy-deception}. Diplomacy dialogues are downloaded from GitHub (all splits). The data are ASCII encoded and newlines are removed. Each message and label is treated as a sample.
This dataset can be downloaded at \url{https://huggingface.co/datasets/diplomacy_detection } under unknown license.

\textbf{echr-decisions}. We collect facts of cases heard before the European Court of Human Rights from \citet{echr-decisions}. Decisions are downloaded from a public archive. A random sample of 500 decisions is selected from the files. The samples with any violated articles are categorized as ``violation,'' while the rest are categorized as ``no violation.''
This dataset can be downloaded at \url{https://paperswithcode.com/dataset/echr} under unknown license.

\textbf{essay-scoring}. We collect essays from students from \citet{essay-scoring}. Essays are downloaded from a GitHub repository. Only essays from set 5 are considered. Essays with a score of at least 3 are categorized as good essays, while essays with a score less than 3 are bad essays.
This dataset can be downloaded at \url{https://www.kaggle.com/c/asap-aes} under unknown license.

\textbf{fake-news}. We collect fake and legitimate news from \citet{fake-news}. Fake news articles are downloaded from the author's website. Full articles are treated as text snippets.
This dataset can be downloaded at \url{http://web.eecs.umich.edu/~mihalcea/downloads.html\#FakeNews} under CC-BY-4.0.

\textbf{fomc-speeches}. We collect Federal Open Market Committee (FOMC) speeches from 1996-2020, which describe Federal Reserve policy from \citet{fomc-speeches}. Fed speeches are downloaded from Kaggle. The macro indicator data are merged in on the year and month. Full speech text is split by paragraph and categorized by speaker, year, and macroeconomic indicator.
This dataset can be downloaded at \url{https://www.kaggle.com/datasets/natanm/federal-reserve-governors-speeches-1996-2020} under unknown license.

\textbf{genius-lyrics}. We collect lyrics collected from Genius.com before 2020 from \citet{genius-lyrics}. Genius lyrics are downloaded from Google Drive. The lyrics are merged with song metadata and treated as samples. We categorize lyrics by hand-selecting popular artists, common genres, time periods, and view counts (over 1M views is high, 500k-1M is medium).
This dataset can be downloaded at \url{https://www.cs.cornell.edu/~arb/data/genius-expertise/} under unknown license.

\textbf{happy-moments}. We collect self-reported happy moments and demographic characteristics from \citet{happy-moments}. The HappyDB dataset is downloaded from the official GitHub repository. Demographic data is cleaned and merged into happy moments. Happy moment descriptions are treated as samples and are categorized by type of happy moment, country of origin, and other demographic features.
This dataset can be downloaded at \url{https://github.com/megagonlabs/HappyDB} under unknown license.

\textbf{huff-post-headlines}. We collect headlines from the news outlet Huffington Post from \citet{huff-post-headlines} and \citet{huff-post-headlines2}. Huffington Post headlines are downloaded from Kaggle. The short description of each article is treated as a sample and tokenized at the sentence level.
This dataset can be downloaded at \url{https://rishabhmisra.github.io/publications/} under CC-BY-4.0.

\textbf{immigration-speeches}. We collect congressional and presidential speeches that mention immigration from 1880 to the present from \citet{immigration-speeches}. Immigration speeches are downloaded from the replication package. The speech text is preprocessed to remove extraneous spaces. We engineer features corresponding to time periods, well-known speakers, other significant time periods, the racial group under discussion, and the geographic area within the United States.
This dataset can be downloaded at \url{https://github.com/dallascard/us-immigration-speeches/releases}.

\textbf{kickstarter}. We collect names of startups on kickstarter.com from \citet{kickstarter}. We download a 2018 crawl from Kickstarter from Kaggle. The project name is treated as the text sample.
This dataset can be downloaded at \url{https://www.kaggle.com/datasets/kemical/kickstarter-projects?select=ks-projects-201612.csv} under CC BY-NC-SA 4.0.

\textbf{microedit-humor}. We collect funny sentences generated by making one-word edits to normal statements from \citet{microedit-humor}. The Microedit dataset is downloaded from the author's website. We make the relevant edit to each text sample and treat the edited text sample as the data point. We bin the mean annotator grade into 4 and denote each as unfunny, neutral, funny, and very funny, respectively.
This dataset can be downloaded at \url{https://paperswithcode.com/dataset/humicroedit}.

\textbf{mnli}. We collect a collection of sentence pairs annotated with textual entailment information from a range of genres from \citet{mnli}. The MNLI corpus is downloaded from the official website. We treat the premise and hypothesis as text samples.
This dataset can be downloaded from \url{https://cims.nyu.edu/~sbowman/multinli/}, most of which are under the OANC license.

\textbf{monster-jobs}. We collect American job postings on monster.com. Jobs on Monster.com are downloaded from Kaggle. Job descriptions are treated as samples and split at the paragraph and sentence level. We keep and categorize jobs from seventeen large cities.
This dataset can be downloaded from \url{https://www.kaggle.com/datasets/PromptCloudHQ/us-jobs-on-monstercom} under CC BY-SA 4.0 .

\textbf{movie-tmdb}. We collect movie plot summaries from TMDB from \citet{movie-tmdb}. TMDB movie overviews are downloaded from Kaggle. We keep only English movies and bin popularity by deciles. The top decile is considered ``hits,'' the 70-80th percentiles are considered ``average,'' and the 30-40th percentiles are considered ``bad.''
This dataset can be downloaded from \url{https://www.kaggle.com/datasets/tmdb/tmdb-movie-metadata21}. 

\textbf{movie-wiki}. We collect movie plot summaries collected from Wikipedia from \citet{movie-wiki}. Wikipedia movie summaries are downloaded from Kaggle.
This dataset can be downloaded from \url{https://www.kaggle.com/datasets/jrobischon/wikipedia-movie-plots} under CC BY-SA 4.0.

\textbf{news-popularity}. We collect news headlines posted on social media platforms from \citet{news-popularity}. Headlines are downloaded from a reproduction package. The headline and title text are cleaned, and the title is treated as the text sample. The 100 most positive and negative or popular and unpopular articles on each topic are used as distributions.
This dataset can be downloaded from \url{https://archive.ics.uci.edu/ml/datasets/News+Popularity+in+Multiple+Social+Media+Platforms}.

\textbf{nli-benchmarks}. We collect training examples from various natural language inference (NLI) datasets from \citet{nli-benchmarks}. NLI benchmarks are downloaded from a public collection on Google Drive. We examine the premise and hypothesis separately as samples.
This dataset can be downloaded from \url{https://github.com/alisawuffles/wanli}.

\textbf{npt-conferences}. We collect Non-Proliferation of Nuclear Weapons (NPT) conference transcripts from \citet{npt-conferences}. NPT conference notes are extracted from the accompanying replication package. Text is split by paragraph, and only paragraphs longer than 50 characters are preserved. Text is split into three time ranges: pre-2008, 2008-2012, and post-2012.
This dataset can be downloaded from \url{https://journals.sagepub.com/doi/full/10.1177/0022343320960523}.

\textbf{open-deception}. We collect arbitrary lies and truths from any domain generated by crowdworkers from \citet{open-deception}. Open domain lies are downloaded from the public dataset and lie texts are split into lies and truths.
This dataset can be downloaded from \url{https://web.eecs.umich.edu/~mihalcea/downloads.html#OpenDeception}.

\textbf{open-review}. We collect submissions to ICLR, a machine learning conference from 2018 to 2021. Open review abstracts are accessed via the openreview API. We query for abstracts from the 2018-2021 ICLR blind submissions. Abstracts are classified based on rating: $>=7$ (``great''), 5-6 (``good''), and $<=4$ (``bad'').
This dataset can be downloaded from \url{https://openreview.net/}. 

\textbf{parenting-subreddits}. We collect posts from various parenting-related subreddits, which are text-based forums on the site Reddit from \citet{parenting-subreddits}. Posts from various subreddits are downloaded from the paper's GitHub repository. We clean the text and split the posts according to the topic(s) each post is tagged with.
This dataset can be downloaded from \url{https://github.com/SALT-NLP/Parenting_OnlineUsage}.

\textbf{poetry}. We collect poems from PoetryFoundation.com from \citet{poetry}. Poems are downloaded from a 2019 scrape of the PoetryFoundation website from Kaggle. The text is cleaned and split according to subject tags and authorship.
This dataset can be downloaded from \url{ https://www.kaggle.com/datasets/tgdivy/poetry-foundation-poems} under GNU Affero General Public License.

\textbf{political-ads}. We collect political ads observed by Facebook users from \citet{political-ads}. Ads are downloaded from the Ad Observer website, which maintains an aggregate of all collected ads. We extract targeting metadata from the targeting field and define splits according to age, gender, location, interests, time, and political lean.
This dataset can be downloaded from \url{https://adobserver.org/ad-database/}. 

\textbf{qqp}. We collect questions from Quora.com from \citet{qqp}.

\textbf{rate-my-prof}. We collect reviews of lecturers from RateMyProfessor.com from \citet{rate-my-prof}. We download a sample of RateMyProfessor.com reviews from an online repo. We clean the text and guess the gender of the reviewed lecturer from the first name using the gender-guesser package. Due to data availability, we consider only male and female names. To improve the quality of the classification, we remove any posts which use pronouns from the opposing sex (e.g. ``him'').
This dataset can be downloaded from \url{https://data.mendeley.com/datasets/fvtfjyvw7d/2} under CC BY 4.0 .

\textbf{radiology-diagnosis}. We collect impressions and medical histories of radiology patients from \citet{radiology-diagnosis}. Radiology diagnoses are downloaded from a GitHub copy of the original task dataset. We parse the metadata to retrieve the diagnostic code, decision type, impression, and patient history. Referencing the associated ICD codes, we convert codes to colloquial diagnoses (e.g. 786.2 denotes cough). We treat the histories and impressions as samples and split them according to diagnosis and level of consensus.

\textbf{reddit-humor}. We collect jokes posted on the Reddit forum r/Jokes, a message board for sharing jokes from \citet{reddit-humor}. Jokes are downloaded from the dev and test splits of the dataset. We clean the text and split the dataset according to whether they are labeled as funny.
This dataset can be downloaded from \url{https://github.com/orionw/rJokesData} under Reddit License and Terms of Service, and users must follow the Reddit User Agreement and Privacy Policy, as well as remove any posts if asked to by the original user.

\textbf{reddit-stress}. We collect stress-related posts on Reddit from \citet{reddit-stress}. We split the post text based on which subreddit they are posted on (related to PTSD, anxiety, or stress generally).
Reddit posts are downloaded from \url{https://github.com/gillian850413/Insight_Stress_Analysis}, and we recommend following the Reddit User Agreement and Privacy Policy, as well as remove any posts if asked to by the original user.

\textbf{reuters-authorship}. We collect articles from various Reuters authors from \citet{reuters-authorship}. The articles are split according to the author. Reuters articles are downloaded from the UCI repository \url{https://archive.ics.uci.edu/ml/datasets/Reuter_50_50}.

\textbf{riddles}. We generated several riddles. The 3000 most common English words are manually copied from a website. Words with between 5 and 8 characters are kept. We create two popular riddles. First, we split words based on whether they have a duplicate character. We exclude any words with multiple ``doubles'' or more than 2 of any character. Second, we split words based on whether they have the letter T.

\textbf{scotus-cases}. We collect facts from cases heard by the Supreme Court of the United States (SCOTUS) from \citet{scotus-cases}. Supreme Court cases are downloaded from a GitHub repository. We identify state/federal parties by manually defining keywords. We split based on the winning party, the identity of each party, and the type of decision. We then define several time periods and relevant political eras and split decisions accordingly. Finally, we split according to the ruling's policy area and how it changes over time.
The dataset can be downloaded from \url{https://paperswithcode.com/paper/justice-a-benchmark-dataset-for-supreme-court} under CC-BY-SA.

\textbf{short-answer-scoring}. We collect short answers from students from \citet{short-answer-scoring}. Short answers are downloaded from a GitHub mirror of the dataset. We consider only responses to essay set 1. The two scores are averaged and binned into good ($>=$ 2.5), medium (1.5-2.5), and bad ($<$1.5).
The dataset can be downloaded from \url{https://www.kaggle.com/c/asap-sas}. 

\textbf{snli}. We collect a collection of sentence pairs annotated with textual entailment information from images from \citet{snli}.
The dataset can be downloaded from \url{https://nlp.stanford.edu/projects/snli/} under CC BY-SA 4.0.

\textbf{squad-v2}. We collect reading comprehension questions crowdsourced from Wikipedia articles from \citet{squad-v2}.
The dataset can be downloaded from \url{https://rajpurkar.github.io/SQuAD-explorer/} under CC BY-SA 4.0.

\textbf{stock-news}. We collect top news headlines on Reddit, an online message board from \citet{stock-news}. Headlines are downloaded from a GitHub mirror. We clean the text and divide the samples based on whether the DOW rose or fell that day.
The dataset can be downloaded from \url{https://github.com/ShravanChintha/Stock-Market-prediction-using-daily-news-headlines} under Reddit License and Terms of Service, and users must follow the Reddit User Agreement and Privacy Policy, as well as remove any posts if asked to by the original user.

\textbf{suicide-notes}. We collect posts from r/SuicideWatch and r/depression, two forums on Reddit from\citet{suicide-notes}. The post title and body are combined to form the text samples. Samples are split based on whether they were posted in a suicide-related Subreddit.
The dataset can be downloaded from a github: \url{https://github.com/hesamuel/goodbye_world},
under Reddit License and Terms of Service, and users must follow the Reddit User Agreement and Privacy Policy, as well as remove any posts if asked to by the original user.

\textbf{times-india-headlines}. We collect headlines from Times of India news from \citet{times-india-headlines}. Headlines are downloaded from a Dataverse mirror. We use the first 1000 headlines in each year as samples.
The dataset can be downloaded from \url{https://www.kaggle.com/datasets/therohk/india-headlines-news-dataset} under CC0 Public Domain.

\textbf{trial-deception}. We collect testimonies from witnesses in real trials from \citet{trial-deception}. Trial testimonies are downloaded from the author's website. The testimonies are divided based on whether they are considered truthful.
The dataset can be downloaded from  \url{https://web.eecs.umich.edu/~mihalcea/downloads.html#RealLifeDeception}. 

\textbf{un-debates}. We collect speeches from debates at the United Nations from \citet{un-debates}. Debate transcripts are downloaded from the Dataverse reproduction package. Samples are divided based on the country and year of the snippet. First, we isolate samples from Russia, China, and the United States and specify 3 time periods of interest. Next, we divide all samples by the decade. Finally, we create distributions for 19 countries of interest.
The dataset can be downloaded from \url{https://doi.org/10.7910/DVN/0TJX8Y} under CC0 1.0 .

\textbf{unhealthy-conversations}. We collect expert-annotated unhealthy conversations from \citet{unhealthy-conversations}. Conversation transcripts are downloaded from the official GitHub repository. For each annotated attribute, we split the dataset based on whether that form of unhealthy conversation is present in the sample.
The dataset can be downloaded from \url{https://github.com/conversationai/unhealthy-conversations} under CC BY-NC-SA 4.0.

\textbf{urban-dictionary}. We collect definitions from UrbanDictionary.com, a crowdsourced English dictionary from \citet{urban-dictionary}. Urban Dictionary entries are downloaded from Kaggle. Definitions are split into groups representing the top 1, 5, and 10 percent of definitions ranked by both upvotes and downvotes; we sample 10,000 from each and create a control distribution by randomly sampling 10,000 definitions from all entries.
The dataset can be downloaded from \url{https://www.kaggle.com/therohk/urban-dictionary-words-dataset} under CC0 Public Domain. 

\textbf{wikitext}. We collect text snippets from Wikipedia from \citet{wikitext}. The Wikipedia snippets are loaded from HuggingFace. We remove any samples that are empty or start with '=' (which represent headings); samples are tokenized at the sentence level and used for clustering.
The dataset can be downloaded from \url{https://huggingface.co/datasets/wikitext} under CC BY-SA 3.0.

\textbf{yc$-$startups}. We collect descriptions of companies that were part of the Y Combinator startup incubator from \citet{yc-startups}. YCombinator company descriptions are downloaded from a 2022 scrape on GitHub. Only companies with long descriptions are preserved. Companies are split according to founder characteristics, year, ``top company'' designation, operating status, and location.
The dataset can be downloaded from \url{https://www.kaggle.com/datasets/benhamner/y-combinator-companies}. 
 \label{app:dataset-descriptions}

%%%%%%%%%%%%%%%%%%%%%%%%%%%%%%%%%%%%%%%%%%%%%%%%%%%%%%%%%%%%

\end{document}